\documentclass[10pt,journal]{IEEEtran}
\usepackage{amsmath}
\usepackage{threeparttable}
\usepackage{amsfonts}
\usepackage{multirow}
\usepackage{booktabs}
\usepackage{amsthm}
\usepackage[noend]{algpseudocode}
\usepackage{algorithmicx,algorithm}
\usepackage{subfigure}
\usepackage{graphicx}
\usepackage{tabularx}
\usepackage{moresize}
\usepackage[numbers,sort&compress]{natbib}
\ifCLASSINFOpdf
\else
\fi

\begin{document}
%
\title{A Brief Review of Real-World Color Image Denoising}

\author{Zhaoming~Kong, Xiaowei~Yang ~\IEEEmembership{}
\IEEEcompsocitemizethanks{\IEEEcompsocthanksitem Zhaoming Kong (E-mail: kong.zm@mail.scut.edu.cn) and Xiaowei Yang (xwyang@scut.edu.cn) are with the school of Software Engineering, South China University of Technology, Guangdong Province, China.\protect\\}}

\maketitle

\begin{abstract}
Filtering real-world color images is challenging due to the complexity of noise that can not be formulated as a certain distribution. However, the rapid development of camera lens poses greater demands on image denoising in terms of both efficiency and effectiveness. Currently, the most widely accepted framework employs the combination of transform domain techniques and nonlocal similarity characteristics of natural images. Based on this framework, many competitive methods model the correlation of R, G, B channels with pre-defined or adaptively learned transforms. In this chapter, a brief review of related methods and publicly available datasets is presented, moreover, a new dataset that includes more natural outdoor scenes is introduced. Extensive experiments are performed and discussion on visual effect enhancement is included.
\end{abstract}


%
\IEEEpeerreviewmaketitle

\section{Introduction}
\begin{figure}[htbp]
\graphicspath{{Illus_image/nonlocal_similarity/}}
\centering
\subfigure{
\label{Fig4}
\includegraphics[width=1.08in]{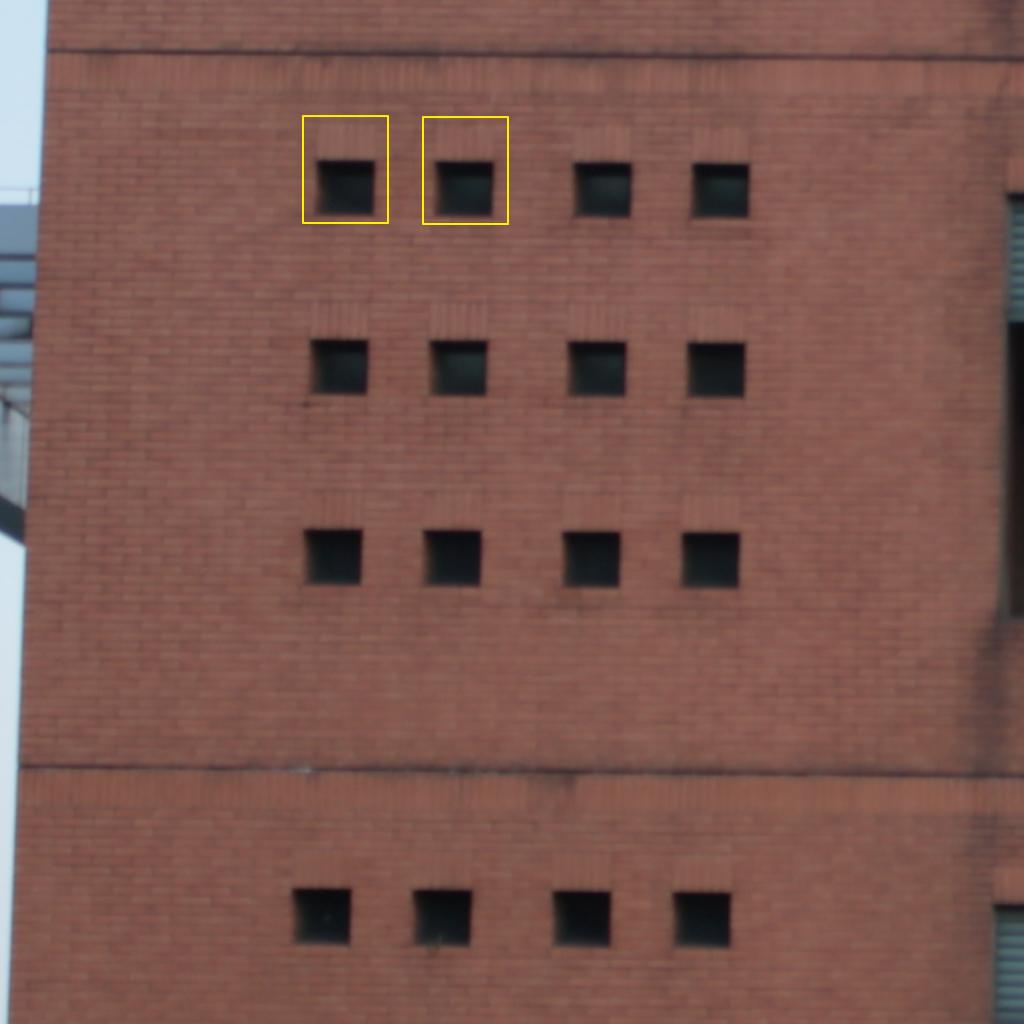}}
\subfigure{
\label{Fig4}
\includegraphics[width=1.08in]{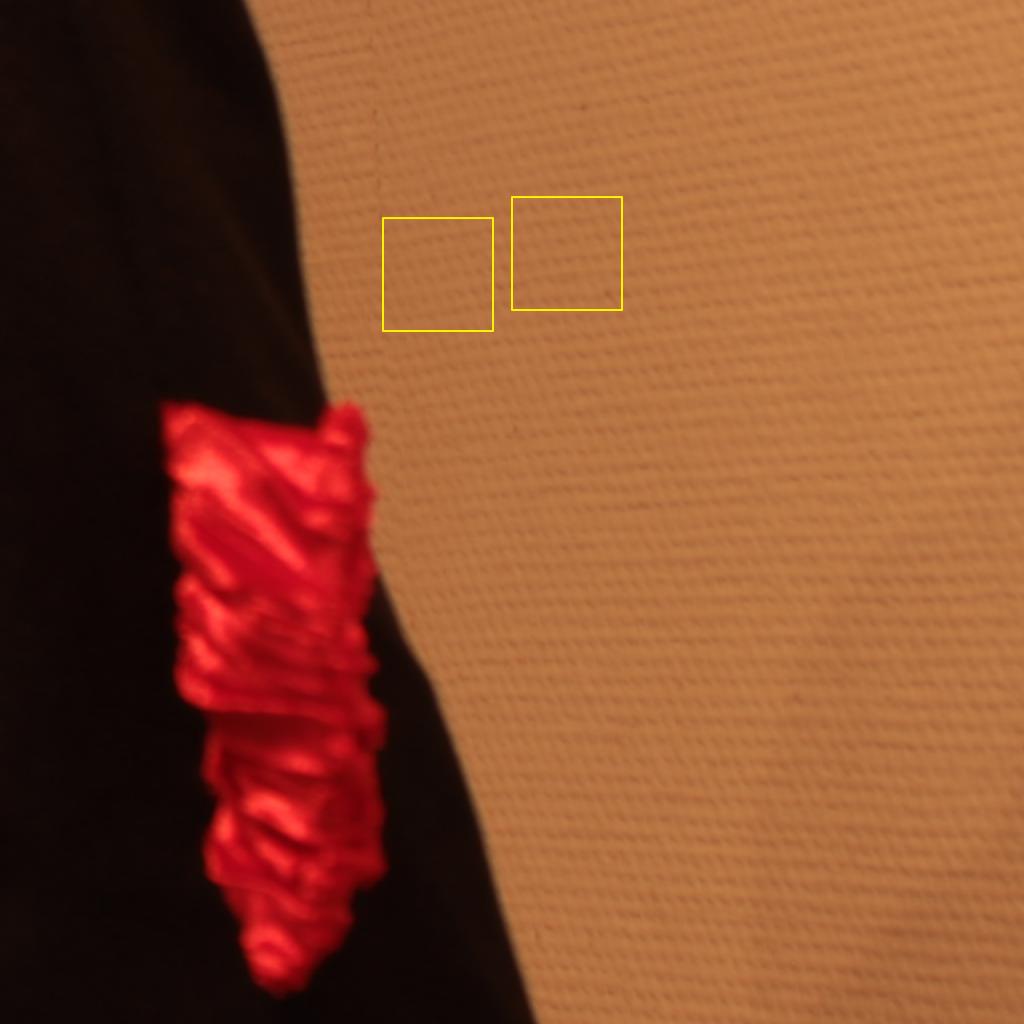}}
\subfigure{
\label{Fig4}
\includegraphics[width=1.08in]{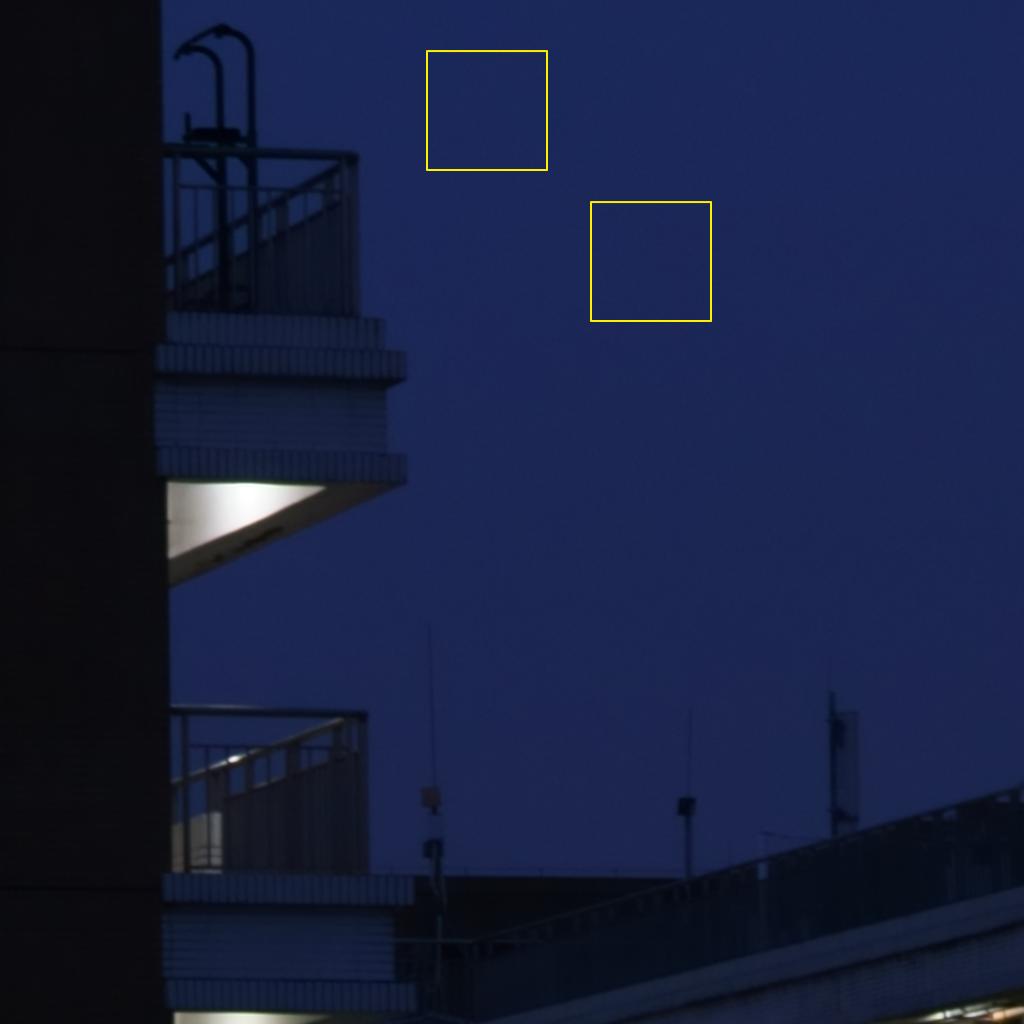}}

\caption{Illustration of non-local similarity feature of natural images.}

\label{Fig_illus_nonlocal_similarity}
\end{figure}

Images are inevitably contaminated by noise during acquisition. According to \cite{Healey1994Radiometric}, the real-world noise of existence in the in-camera process \cite{Nam2016A,Tsin2001Statistical,Kim2012A} is signal dependent and stems mainly from five sources: photon shot, fixed pattern, dark current, readout and quantization noise. With the presence of noise, possible subsequent image processing tasks such as video processing, image analysis and tracking are adversely affected. Therefore, image denoising plays an important role in modern image processing systems.\\
\indent The past few decades witness great achievements in the field of image denoising \cite{Milanfar2013A}. The non-local and similarity feature \cite{Buades2005A} of natural images, as illustrated in Fig. \ref{Fig_illus_nonlocal_similarity}, is mainly employed for designing image denoising methods. Furthermore, transform domain \cite{Yaroslavsky2001Transform} technique is introduced according to the assumption that signal will be sparsely distributed. The combination of non-local similarity property and transform domain technique is a simple and effective framework, which can be roughly divided into three consecutive steps: grouping, collaborative filtering and aggregation. More detailed description is given in Algorithm \ref{nonlocal_transform_framework}. Following similar idea, there are several interconnected criteria to categorize existing methods, and techniques adopted by some representative methods are briefed in Table \ref{Table_technique_generalization}.\\
\begin{table*}[htbp]
\ssmall
\centering
  \caption{Techniques adopted by representative methods. PCA: principal componen analysis, LR: low rank, RN: residual network, TF: tensor factorization, SC: sparse coding, HT: hard-thresholding, WF: Wiener filter.}
    \begin{tabular}{ccccccccccc}
    \toprule
    Method & LSCD \cite{rizkinia2016local}  & LLRT \cite{chang2017hyper}  & DnCNN \cite{Harmeling2012Image} & TNRD \cite{Chen2015On} & MCWNNM \cite{xu2017multi} & GID \cite{xu2018external}  & TWSC \cite{TrilateralXu} & 4DHOSVD1 \cite{rajwade2013image} & CBM3D \cite{Dabov2007Image} & MS-TSVD \cite{Zhaoming}\\
    \midrule
    Techique & Color Line + PCA & LR + TF & RN & External Prior & Weighted LR & External+Internal Prior & Weighted SC & HT + TF & HT+WF & HT + TF \\
    \bottomrule
    \end{tabular}%
  \label{Table_technique_generalization}%
\end{table*}%

\indent First, based on the choice of transform, numerous methods are devised. Early works are from bilateral filters \cite{BilateralTomasi}, wavelet \cite{Chang2002Adaptive} and curvelet \cite{Starck2002The} transforms. Then the well-known color block-matching 3D (CBM3D) \cite{Dabov2007Image} combines discrete cosine transform (DCT) with opponent color mode transform, and produces state-of-the-art performance. Instead of using predefined transforms, some recently proposed methods adopt adaptively learned transforms. \cite{Gu2014Weighted} uses singular value decomposition (SVD) and designs a weighted nuclear norm (WNNM) strategy. \cite{rajwade2013image} uses the idea of tensor \cite{Kolda2009Tensor} and trains corresponding mode transforms with 4D higher-order singular value decomposition (4DHOSVD). Most recently, \cite{Zhaoming} utilizes tensor-SVD \cite{Kilmer2013Third} and characterizes color image with block circulant representation \cite{Mazancourt1983The}.\\
\indent The second difference lies in the modeling of noise. Most of existing methods \cite{Chang2002Adaptive,Starck2002The,Dabov2007Image,Gu2014Weighted,rajwade2013image,Zhaoming,Lebrun2013A,EPLLZoran,Xu2015Patch} consider the simplest additive white Gaussian noise (AWGN), and aim to recover the clean image $\mathcal{X}$ from its noisy observation $\mathcal{Y} = \mathcal{X} + \mathcal{N}$, where $\mathcal{N}$ is AWGN. Typically, $l_2$-norm is combined with different regularization to enforce sparsity in the transform domain. Efforts beyond AWGN model are also made. \cite{Zhang2008Wavelets} and \cite{Salmon2014Poisson} propose Poisson noise reduction models, \cite{Luisier2011Image} and \cite{Le2014An} consider mixed Poisson-Gaussian noise, while \cite{Jiang2014Mixed} and \cite{Xu2016patch} handle mixed Gaussian and impulsive noise. Besides, there are also some methods \cite{Liu2008Automatic,Lebrun2015The,Lebrun2015Multiscale} and well-known software tool box Neat Image (NI) \footnote{Neatlab ABSoft. https://ni.neatvideo.com/home} developed for real-world noisy image denoising.\\
\indent Last but not least, many methods vary according to how correlation among R, G, B channels are characterized. \cite{Dai2013Multichannel} proposes a  multichannel nonlocal fusion (MNLF) approach, which constructs and fuses multiple NLM across three channels together. \cite{tu2014collaborative} introduces color line to model the correlation among neighbouring pixels as well as among RGB channels, and \cite{xu2017multi} extends WNNM to handle color image by using a weight matrix to balance noise in the RGB channel. Moreover, some methods apply a de-correlation transform to the RGB color space so that grayscale image filtering techniques can be directly utilized. \cite{HueBen} proposes to suppress noise in the hue-saturation-value space due to the good continuation feature of hue space. The representative CBM3D first transforms the RGB channel into chrominance-luminance color space, then performs noise removal with BM3D \cite{Dabov2007Image} on every component in the new space separately.\\
\indent In order to verify the effectiveness and efficiency of above methods, several real-world color image datasets \cite{Anaya2014RENOIR,Nam2016A,Pl2017Benchmarking,DatasetXu} are constructed, where each scene includes noisy and "ground-truth" image pairs. According to \cite{DatasetXu}, a reasonable idea to obtain "ground-truth" image is to capture the same and unchanged scene for many times and compute their mean image, because for each pixel the noise is assumed to be larger or smaller than 0, and thus can be greatly suppressed by sampling the same pixel for many times. A major problem with existing methods is that they mainly consist of indoor scenes and artificial objects. To comprehensively understand the effectiveness of competing methods, a new dataset that includes many natural outdoor scenes is constructed. Also, compared with indoor scenes, the lighting conditions and illuminations of outdoor scenes are more complicated, thus making denoising more challenging.\\
\indent In this paper, background knowledge is introduced in Section II. A brief description of existing and newly constructed datasets is included in Section III. Extensive experiments and discussion are presented in Section IV. Conclusion is given in Section V.
\begin{algorithm}[ht]
\caption{Nonlocal and transform-domain framework} 
{\bf Input:} Color image $\mathcal{A}$, transform $T$, inverse transform $T^{-1}$, patch size $ps$, local search window size $SR$, number of similar patches $K$, pixels between two adjacent reference patches $N_{step}$.\\
{\bf Output:} Filtered image $\mathcal{A}_{f}$.\\
{\bf Step 1} (Grouping): For a given reference patch, use a certain criteria (often nearest neighbourhood) to stack $K$ similar patches in a group $\mathcal{G}$.\\
{\bf Step 2} (Collaborative filtering):\\
 \hspace*{0.2in}(1) Apply transform T to $\mathcal{G}$ and obtain $\mathcal{G}_{T}$.\\
 \hspace*{0.2in}(2) Apply threshold technique (often low-rank or hard-threshold) to $\mathcal{G}_{T}$ in the transform domain..\\
 \hspace*{0.2in}(3) Apply inverse transform $T^{-1}$ to thresholded $\mathcal{G}_{T}$ and obtain filtered group $\mathcal{G}_{filtered}$.\\
{\bf Step 3} (Aggregation): (Often averagely) Write back all image patches in $\mathcal{G}_{filtered}$ to their original locations and obtain filtered image $\mathcal{A}_{f}$.
\label{nonlocal_transform_framework}
\end{algorithm}
\section{Background Knowledge}

\subsection{Notations and Preliminaries}
In this paper, tensors are denoted by calligraphic letters, e.g., $\mathcal{A}$, matrices by boldface capital letters, e.g. $\mathbf{A}$; vectors are denoted by lowercase boldface letters, e.g., $\mathbf{a}$. The $i$th entry of a vector $\mathbf{a}$ is denoted by $a_i$, element $(i,j)$ of a matrix $\mathbf{A}$ by $a_{ij}$, and element $(i,j,k)$ of a third-order tensor $\mathcal{A}$ by $\mathcal{A}_{ijk}$. An $N$th-order tensor is denoted as $\mathcal{A} \in \mathbb{R}^{I_1\times I_2\times\cdots\times I_N}$. The $n$-mode product of a tensor $\mathcal{A}$ by a matrix $\mathbf{U}\in \mathrm{R}^{P_n\times I_n}$, denoted by $\mathcal{A}\times _n\mathbf{U}$ is also a tensor. The Frobenius norm of a tensor $\mathcal{A} \in \mathbb{R}^{I_1\times I_2\times\cdots\times I_N}$ is defined as $\|\mathcal{A}\|_F = \sqrt{\langle\mathcal{A},\mathcal{A}\rangle}$. The mode-$n$ matricization or unfolding of a $\mathcal{A}$, denoted by $\mathcal{X}_n$, maps tensor elements $(i_1,i_2,\ldots,i_N)$ to matrix element $(i_n,j)$ where $j=1+\sum_{k=1,k\neq n}^{N}(i_k-1)J_k$, with $J_k = \prod_{m=1,m\neq n}^{k-1}I_m$.\\
\indent Given a tensor $\mathcal{A}\in \mathbb{R}^{I_1\times I_2\times\cdots\times I_N\times M}$, if $\mathcal{A}$ can be expressed as the product
\begin{equation}\label{kron}
  \mathcal{A} = \mathcal{C} \times _1\mathbf{U}^{(1)^T}\times _2\mathbf{U}^{(2)^T}\times\cdots\times _N\mathbf{U}^{(N)^T}
\end{equation}
\noindent where $\mathcal{C}$ is core tensor and $\mathbf{U}^{(n)}$ represents orthogonal factor matrix, then the matricized version of (\ref{kron}) is
\begin{multline}\label{matricized_kron}
  \mathbf{A}_{(n)} = \mathbf{U}^{(n)^T}\mathcal{C}_{(n)}(\mathbf{U}^{(N)}\otimes\mathbf{U}^{(N-1)}\otimes \\
   \otimes\cdots \mathbf{U}^{(n+1)} \otimes\mathbf{U}^{(n-1)} \otimes\ \cdots \mathbf{U}^{(1)})
\end{multline}
where $\bigotimes$ denotes the Kronecker product.

\subsection{Related methods}
\begin{figure}[htbp]
\graphicspath{{Illus_image/Other/}}
\centering
\subfigure[Vector representation]{
\label{Fig4}
\includegraphics[width=1.4in]{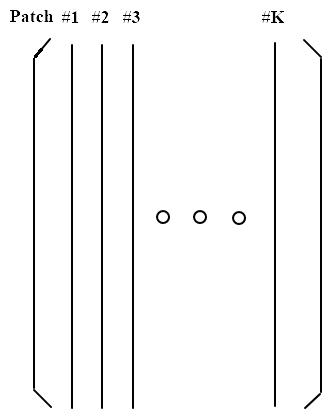}}
\subfigure[Tensor representation]{
\label{Fig4}
\includegraphics[width=1.4in]{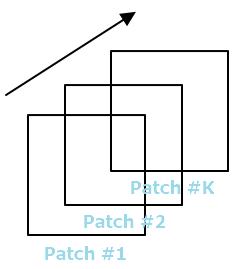}}

\caption{Vector and tensor representation of patches.}

\label{Fig_illus_patch_representation}
\end{figure}

\indent Regardless of various categorization criteria, almost all related methods under the nonlocal and transform domain framework employ the idea of multiway filtering technique \cite{muti2008lower}, and apply different regularization to filter group of patches along each dimension. Vector and tensor are two commonly used representation to model each color image patch, as is illustrated in Fig. \ref{Fig_illus_patch_representation}. Vector representation based methods such as TWSC \cite{TrilateralXu}, MCWNNM \cite{xu2017multi}, LSCD \cite{rizkinia2016local} can benefit from good and vastly-studied statistical properties of singular value decomposition (SVD) or principal component analysis (PCA). However, each vectorized patch can be a lengthy vector and increases computational burden. To alleviate this issue, many recently proposed methods such as 4DHOSVD, LLRT \cite{chang2017hyper} and MS-TSVD exploit tensor representation. In fact, pioneer work is from CBM3D, although tensor is not explicitly stated.
\\
\indent Currently, CBM3D, 4DHOSVD and MS-TSVD demonstrate the most competitive performance in terms of both effectiveness and efficiency, and since tensor can be regarded as an extension of vector to higher dimension, we mainly introduce these three state-of-the-art methods and compare their similarity and difference. More detailed analysis is given in \cite{Zhaoming}.
\\
\indent For a group of patches $\mathcal{G}$, reconstruction problem of both CBM3D and 4DHOSVD can be represented by fourth-order tensor decomposition
\begin{equation}\label{cbm3d_hosvd}
  \mathcal{C} = \mathcal{G} \times _1\mathbf{U}_{row} \times _2\mathbf{U}_{col} \times _3\mathbf{U}_{color} \times _4\mathbf{U}_{group}
\end{equation}
where $\mathbf{U}$ are corresponding mode transform matrices along each dimension of $\mathcal{G}$. For CBM3D, all transforms in (\ref{cbm3d_hosvd}) are predefined, and its opponent color mode transform matrix $\mathbf{U}_{color}$ is
\begin{equation}\label{cbm3d_color_mode_matrix}
\left(
  \begin{array}{ccc}
    1/3 & 1/3 & 1/3 \\
    0.5 & 0 & -0.5 \\
    0.25 & -0.5 & 0.25 \\
  \end{array}
\right)
\end{equation}
Specifically, assume $\mathcal{P}_{new} = \mathcal{P} \times _3{U}_{color}$, then the first slice of $\mathcal{P}_{new}$ in the new color mode can be regarded as luminance channel, while the rest two slices are chrominance channel. CBM3D is very efficient because it does not have to train local transforms, and its grouping step is performed only on the luminance channel. For 4DHOSVD, all mode transform matrices are learned by solving
\begin{equation}\label{optimize_4dhosvd}
\begin{split}
   &  \min \|\mathcal{G} - \mathcal{C} \times _1\mathbf{U}^T_{row} \times _2\mathbf{U}^T_{col} \times _3\mathbf{U}^T_{color} \times _4\mathbf{U}^T_{group}\|_F^2\\
    & s.t  \qquad \mathbf{U}_{row}^T\mathbf{U}_{row} = \mathbf{I},  \qquad \mathbf{U}_{col}^T\mathbf{U}_{col} = \mathbf{I} \\
    & \quad\qquad \mathbf{U}_{color}^T\mathbf{U}_{color} = \mathbf{I}, \qquad \mathbf{U}_{group}^T\mathbf{U}_{group} = \mathbf{I}
\end{split}
\end{equation}
where $\mathbf{I}$ is identity matrix and $\mathbf{U}$ can be obtained via SVD of corresponding unfolding matrix. Different from CBM3D and 4DHOSVD that use direct folding and unfolding of tensor, MS-TSVD adopts the idea of tensor-SVD (t-SVD), and characterizes each color image patch as a block circulant matrix, which can be represented by
\begin{equation}\label{bcirc}
  bcirc(\mathcal{P}) =  \left(
                          \begin{array}{ccc}
                            R & B & G \\
                            G & R & B\\
                            B & G & R \\
                          \end{array}
                        \right)
\end{equation}
\begin{figure}[htbp]
\graphicspath{{Illus_image/Other/}}
\centering

\includegraphics[width=3.38in]{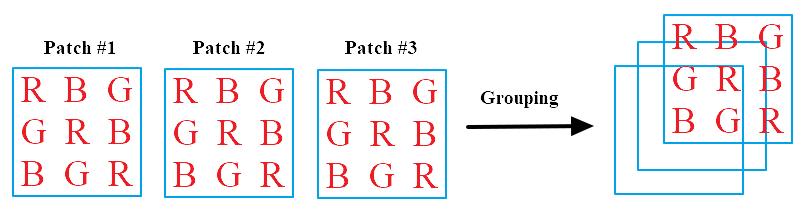}

\caption{Block circulant representation of a group of similar patches.}

\label{Fig_illus_bcirc_group}
\end{figure}

\noindent where $R, G, B$ are three color channels of patch $\mathcal{P}$. Fig. \ref{Fig_illus_bcirc_group} illustrates a group of color image patches that use block circulant representation. Then, according to tensor decomposition in (\ref{kron}), the reconstruction problem of MS-TSVD can be written as
\begin{equation}\label{mstsvd_core_tensor}
  \mathcal{C} = bcirc(\mathcal{G}) \times _1\mathbf{U}_{row} \times _2\mathbf{U}_{col} \times _3\mathbf{U}_{group}
\end{equation}
the factor matrices of (\ref{mstsvd_core_tensor}) can be obtained by solving
\begin{equation}\label{mstsvd_origin_optimize}
\begin{split}
   &  \min \|bcirc(\mathcal{G}) - \mathcal{C} \times _1\mathbf{U}^T_{row} \times _2\mathbf{U}^T_{col} \times _3\mathbf{U}^T_{group}\|_F^2\\
    & s.t \: \mathbf{U}_{row}^T\mathbf{U}_{row} = \mathbf{I}, \: \mathbf{U}_{col}^T\mathbf{U}_{col} = \mathbf{I}, \: \mathbf{U}_{group}^T\mathbf{U}_{group} = \mathbf{I}
\end{split}
\end{equation}
Directly solving (\ref{mstsvd_origin_optimize}) is time-consuming, because it includes the decomposition of a large block circulant matrix. Thus the authors first perform fast Fourier transform (FFT) \cite{Yang2004Fast} along the third mode of $\mathcal{P}$ to convert the block circulant representation into diagonal representation in the Fourier domain. Each patch $\hat{\mathcal{P}}$ in the Fourier domain can be obtained by $\hat{\mathcal{P}} = \mathcal{P} \times _3\mathbf{U}_{FFT}$, where $\mathbf{U}_{FFT}$ is the FFT matrix defined as
 \begin{equation}\label{define_fft}
   \mathbf{U}_{FFT} = \left(
                        \begin{array}{ccc}
                          1 & 1 & 1 \\
                          1 & -0.5-0.8660i & -0.5+0.8660i \\
                          1 & -0.5+0.8660i & -0.5-0.8660i \\
                        \end{array}
                      \right)
 \end{equation}
then problem (\ref{mstsvd_origin_optimize}) can be reformulated as
\begin{equation}\label{optimize_hosvd_bdiag_tensor}
\begin{split}
   &  \min \|\hat{\mathcal{G}} - \hat{\mathcal{C}} \times _1\hat{\mathbf{U}}^T_{row} \times _2\hat{\mathbf{U}}^T_{col} \times _3\hat{\mathbf{U}}^T_{group}\|_F^2\\
    s.t \: &  \hat{\mathbf{U}}_{row}^T \hat{\mathbf{U}}_{row} = \mathbf{I}, \: \hat{\mathbf{U}}_{col}^T\hat{\mathbf{U}}_{col} = \mathbf{I}, \: \hat{\mathbf{U}}_{group}^T\hat{\mathbf{U}}_{group} = \mathbf{I}
\end{split}
\end{equation}
where $\hat{\mathcal{G}} =  \mathcal{G}\times _3\mathbf{U}_{FFT}$. According to \cite{Zhaoming}, $\hat{\mathbf{U}}_{group} = \mathbf{U}_{group}$. Furthermore, from (\ref{define_fft}), two interesting features of FFT and correspondingly of patch $\hat{\mathcal{P}}$ in the Fourier domain can be seen. First, the second and third slices of $\hat{\mathcal{P}}$ are conjugate. Second, the first slice of $\hat{\mathcal{P}}$ in the Fourier domain can be regarded as luminance channel of (\ref{cbm3d_color_mode_matrix}). Therefore, to train the group-wise transform $\mathbf{U}_{group}$ more effectively, an alternative to (\ref{optimize_hosvd_bdiag_tensor}) is using only the luminance channel information.
 Indeed MS-TSVD can be regarded as a generalization of both CBM3D and 4DHOSVD because it uses a predefined color mode transform and applies HOSVD to every new channel in the Fourier domain.
\\
\indent Assume that the number of image pixels is $N$, that the average time to compute similar patches per reference patch is $T_{s}$, that the average number of patches similar to the reference patch is $K$, and that the size of the patch is $p\times p$ ($p\ll K$). According to \cite{rajwade2013image}, the time complexity of 4DHOSVD and CBM3D are $O([T_s+Kp^3+Kp^4]N)$ and $O([T_s+Kp^2logp+p^2KlogK]N)$, respectively. The computational burden of MS-TSVD mainly lies in the PCA transform $O(Kp^4)$ and patch level t-SVD transform $O(Kp^3)$, leading to a total complexity of $O([T_s+Kp^3+Kp^4]N)$. Considering MS-TSVD is a one-step algorithm, it is competitive in terms of efficiency, because in real cases, the input estimate of AWGN should be carefully tuned for best visual effects, and some intermediate results such as grouping index and local PCA can be saved to avoid re-calculation.
\section{Real Datasets}
\begin{figure}[htbp]
\graphicspath{{Illus_image/RENOIR/}}
\centering
\subfigure[CANON S90]{
\label{Fig4}
\includegraphics[width=0.98in]{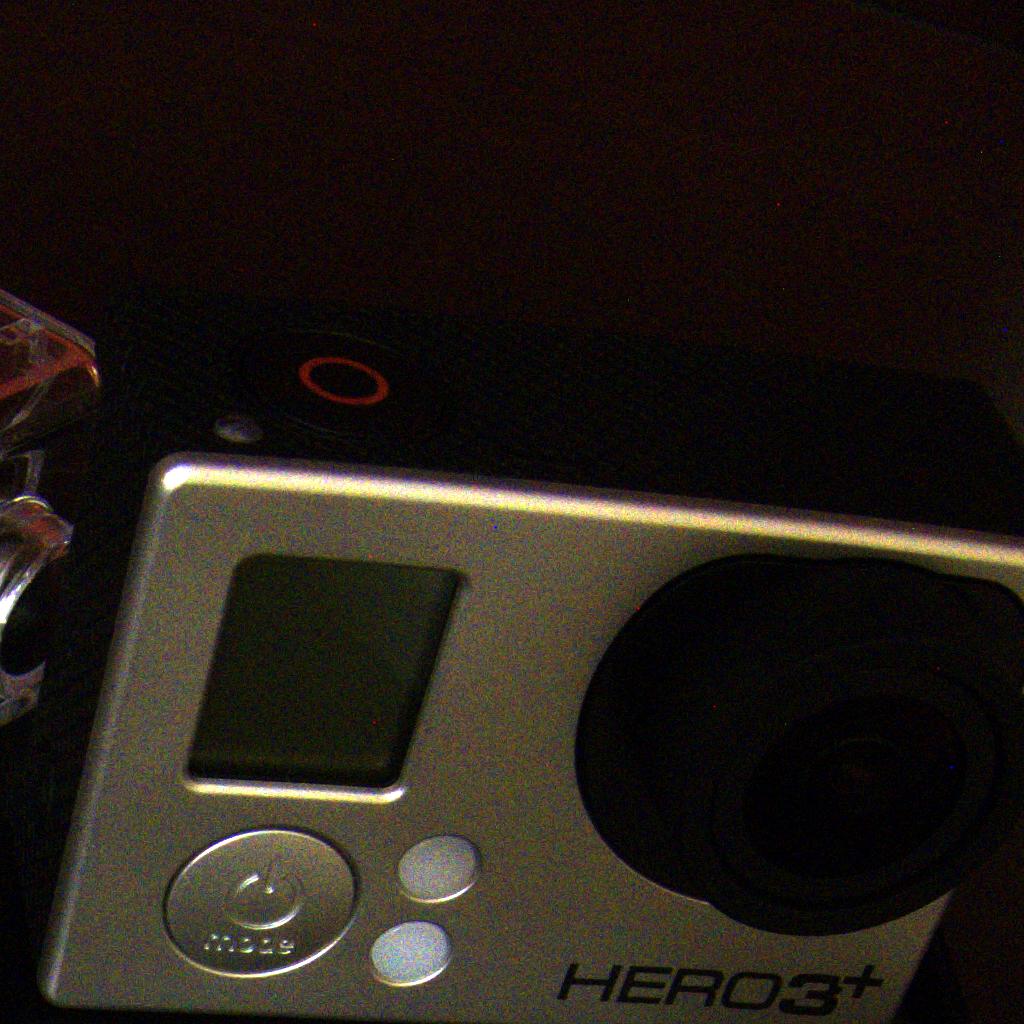}}
\subfigure[CANON T3i]{
\label{Fig4}
\includegraphics[width=0.98in]{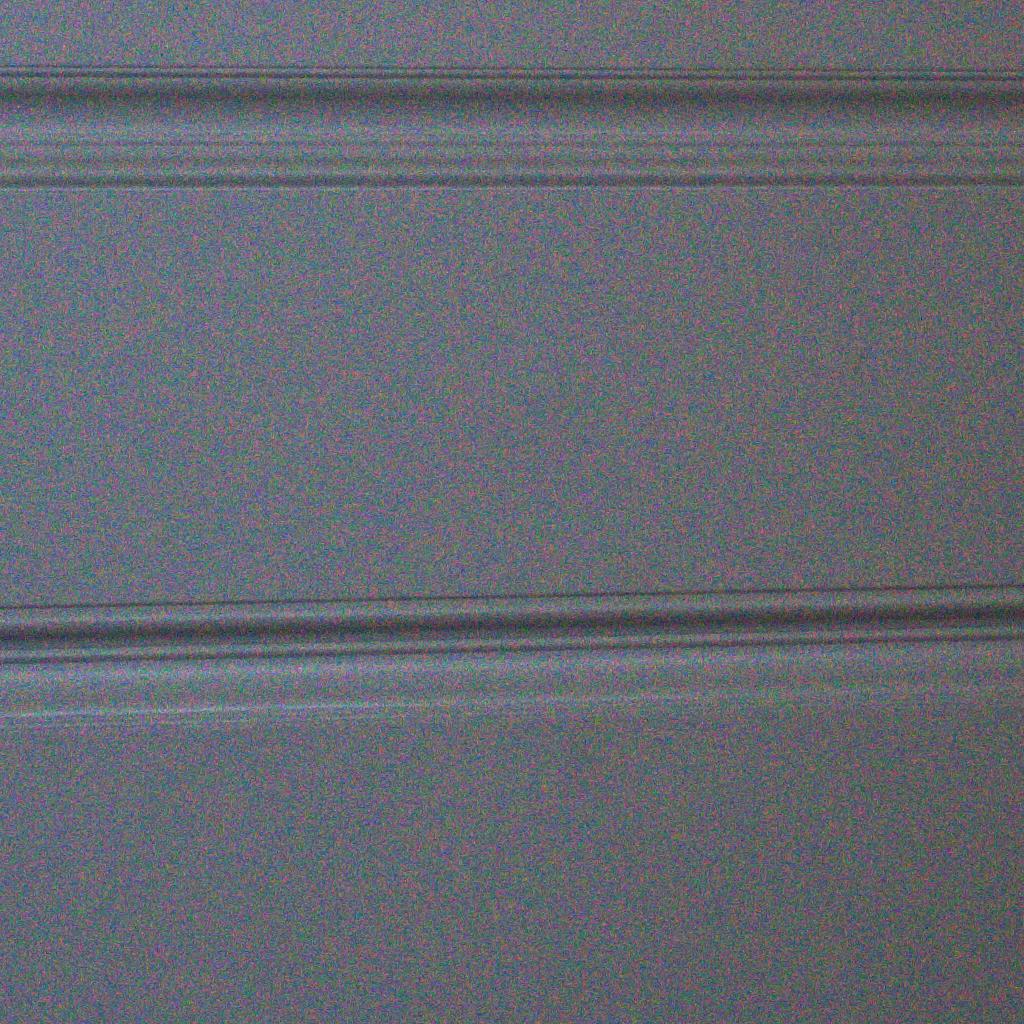}}
\subfigure[XIAOMI Mi3]{
\label{Fig5}
\includegraphics[width=0.98in]{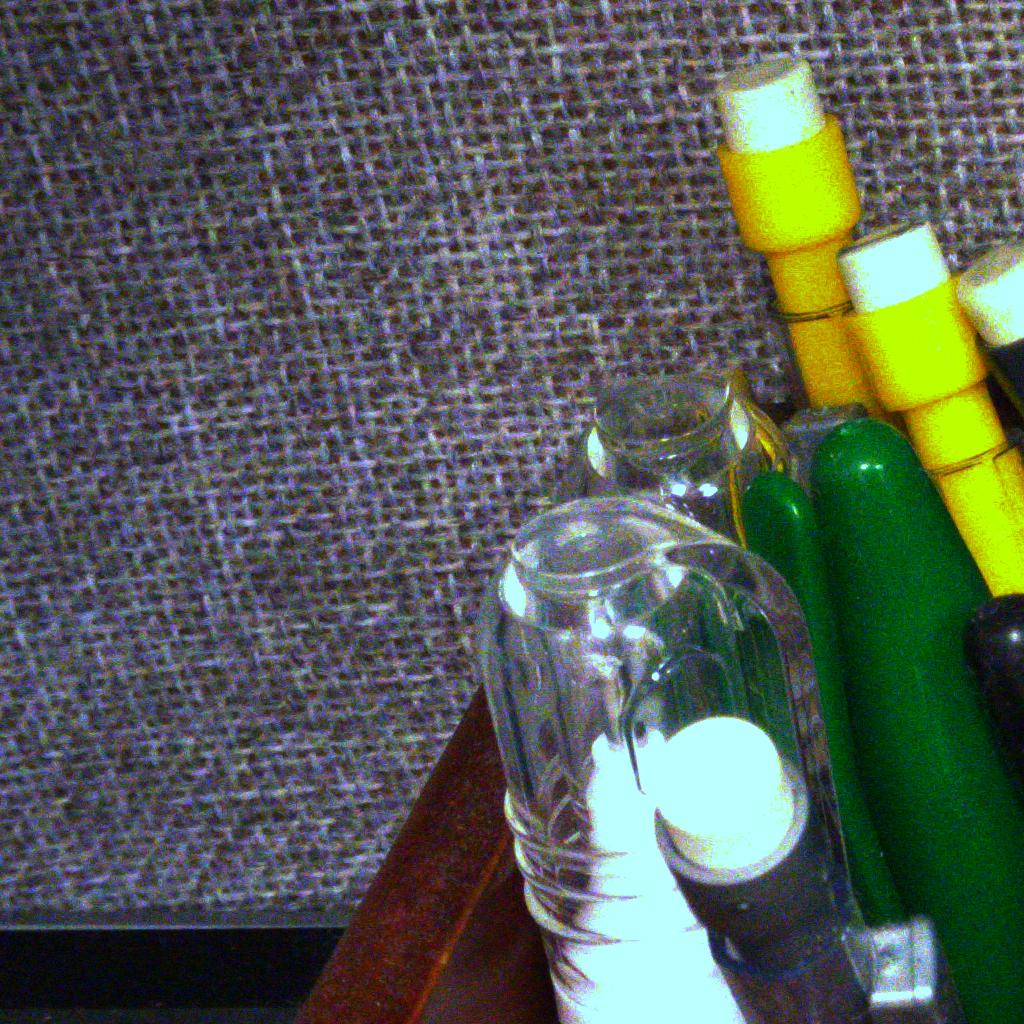}}

\caption{Some sample real noisy images of RENOIR dataset.}

\label{Fig_illus_dataset_renoir}
\end{figure}

\begin{figure}[htbp]
\graphicspath{{Illus_image/Nam/}}
\centering
\subfigure[CANON 5D MARK III]{
\label{Fig4}
\includegraphics[width=0.98in]{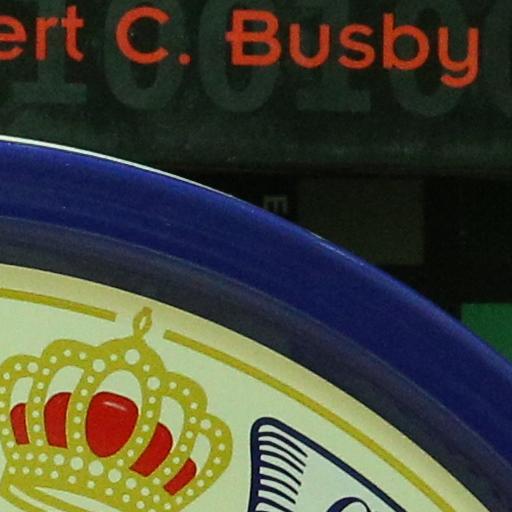}}
\subfigure[NIKON D600]{
\label{Fig4}
\includegraphics[width=0.98in]{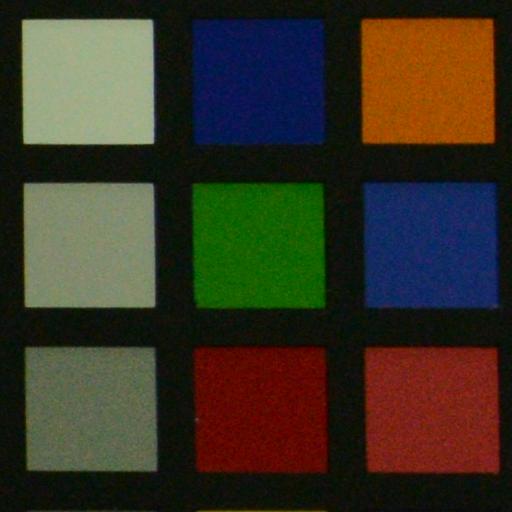}}
\subfigure[NIKON D800]{
\label{Fig5}
\includegraphics[width=0.98in]{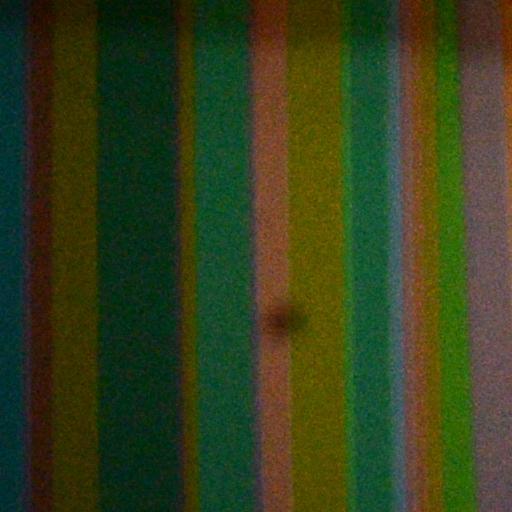}}

\caption{Some sample real noisy images of Nam dataset.}

\label{Fig_illus_dataset_nam}
\end{figure}

\begin{figure}[htbp]
\graphicspath{{Illus_image/DND/}}
\centering
\subfigure[NEXUS 6P]{
\label{Fig4}
\includegraphics[width=0.98in]{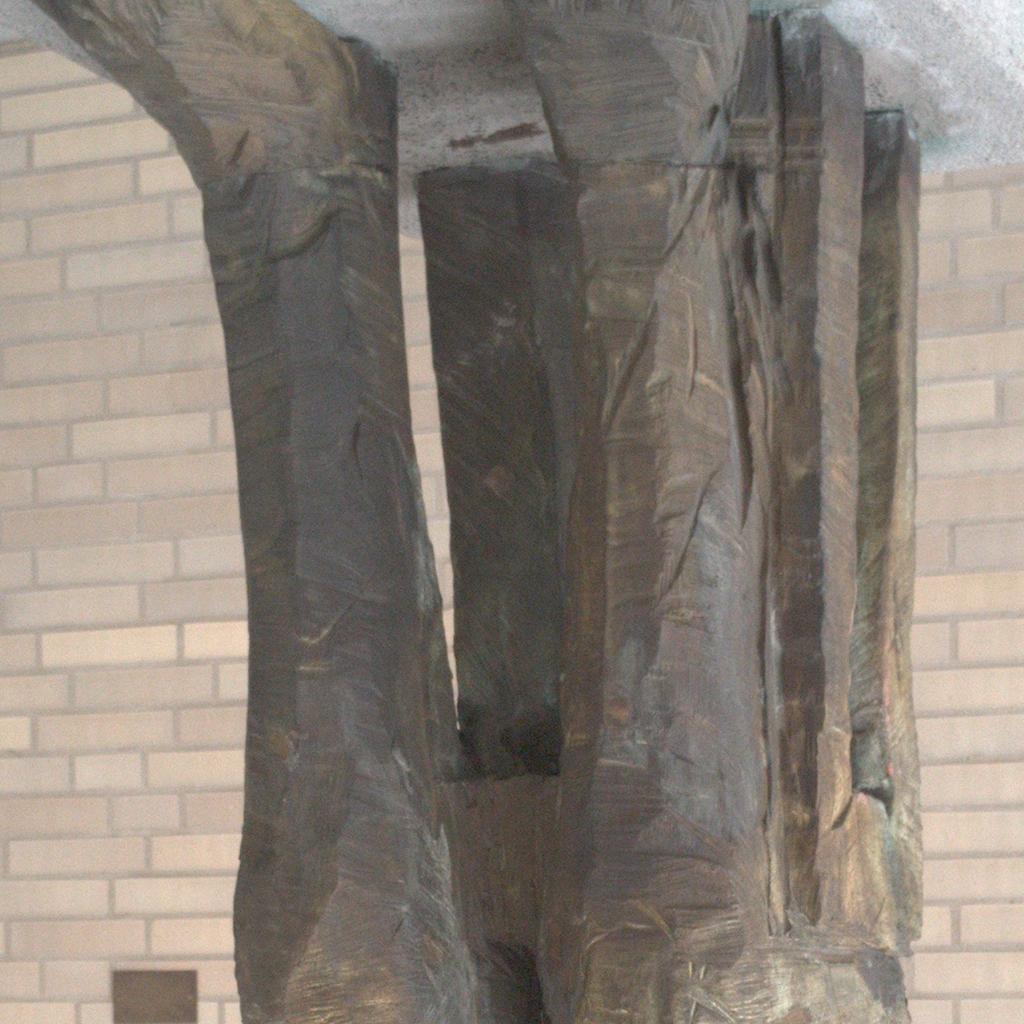}}
\subfigure[OLYMPUS E-M10]{
\label{Fig4}
\includegraphics[width=0.98in]{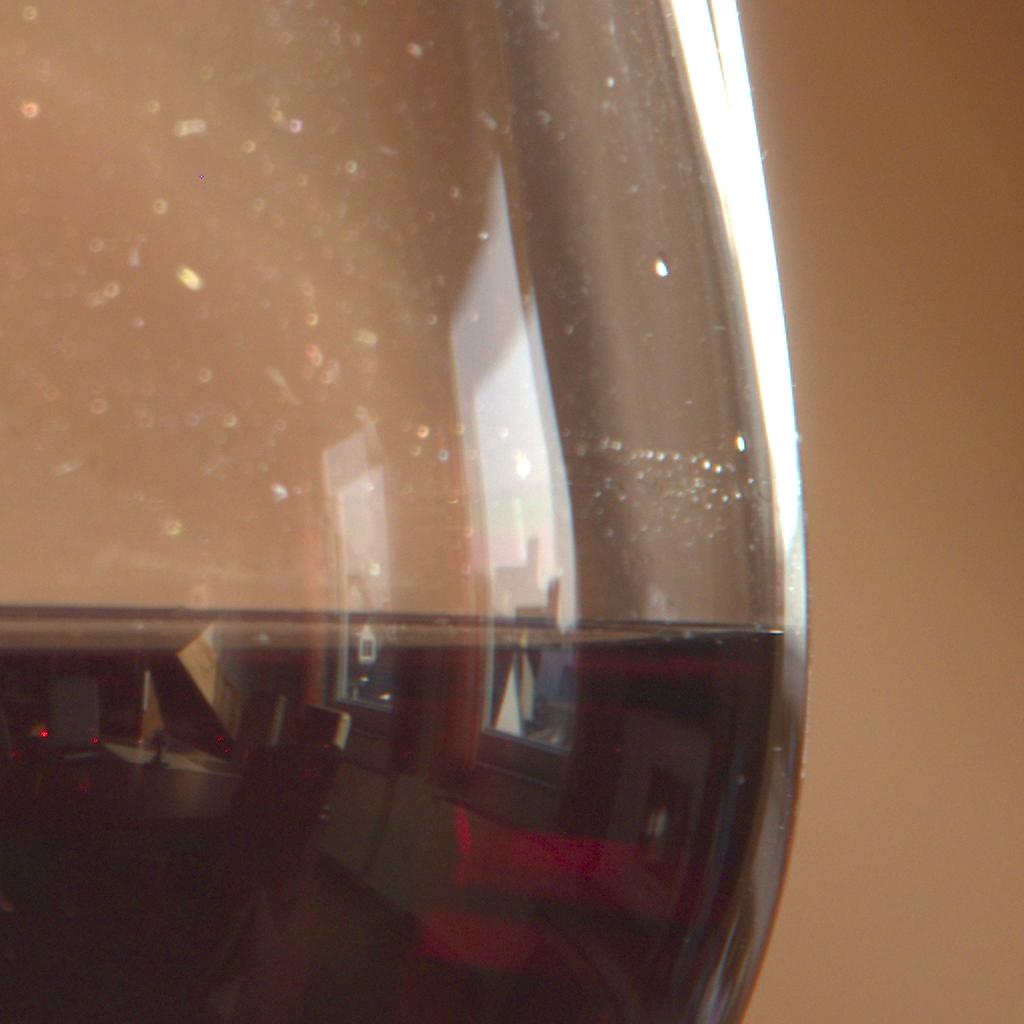}}
\subfigure[SONY A7R]{
\label{Fig5}
\includegraphics[width=0.98in]{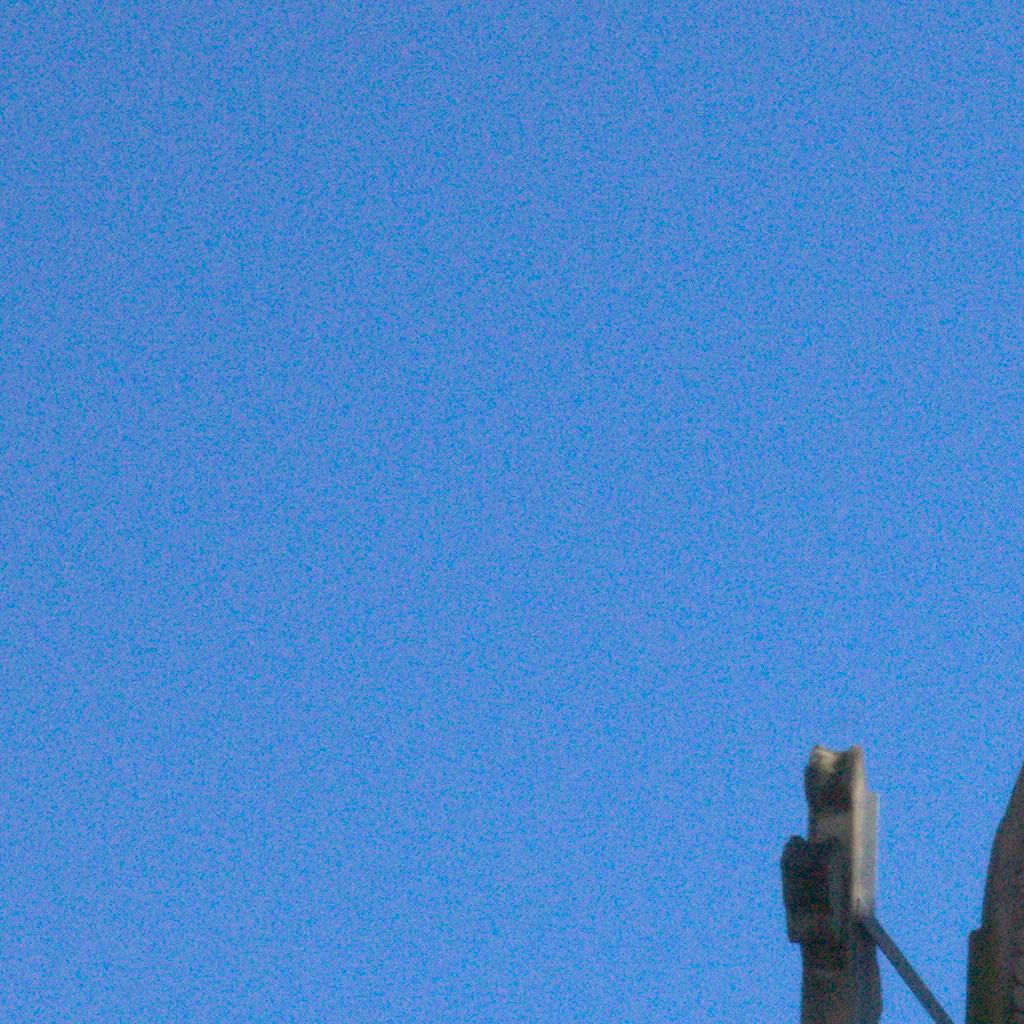}}

\caption{Some sample real noisy images of DND dataset.}

\label{Fig_illus_dataset_dnd}
\end{figure}

\begin{figure}[htbp]
\graphicspath{{Illus_image/Xu/}}
\centering
\subfigure[CANON 5D]{
\label{Fig4}
\includegraphics[width=0.98in]{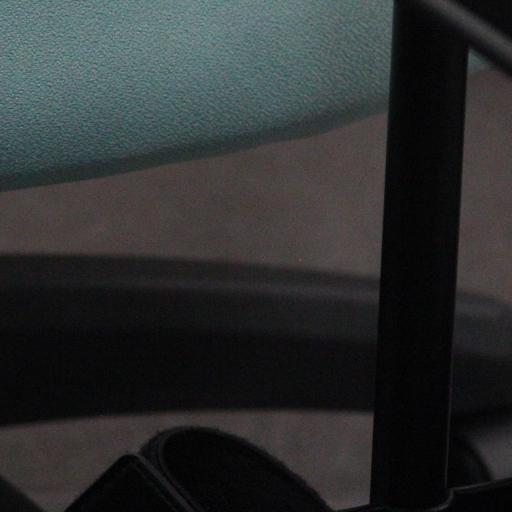}}
\subfigure[CANON 80D]{
\label{Fig4}
\includegraphics[width=0.98in]{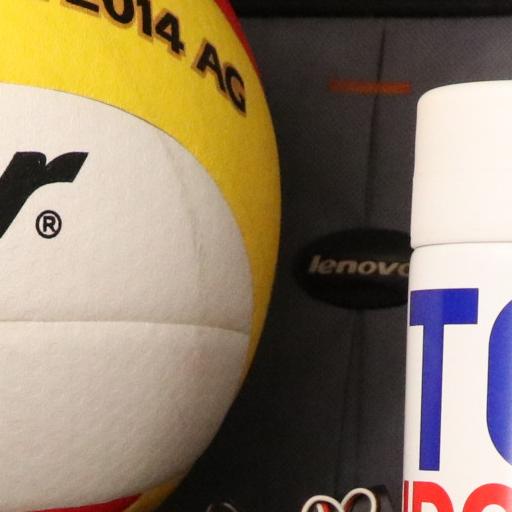}}
\subfigure[NIKON D800]{
\label{Fig5}
\includegraphics[width=0.98in]{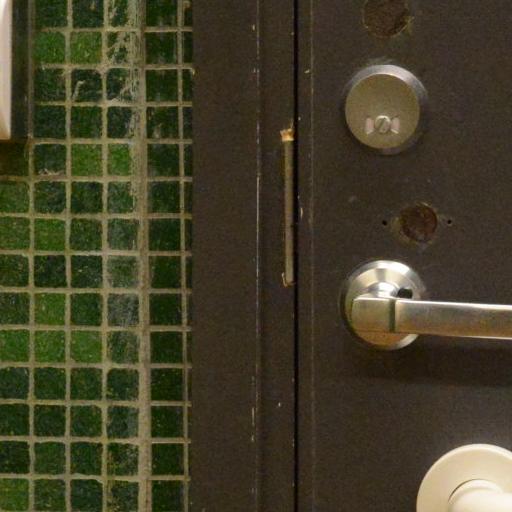}}

\caption{Some sample real noisy images of Xu dataset.}

\label{Fig_illus_dataset_Xu}
\end{figure}

Currently, there are four publicly available datasets used to evaluate denoising methods on real-world noisy images. In this section, these existing real datasets are briefly described and a newly constructed dataset is introduced. Information of all included datasets is given in Table \ref{Table_dataset_description}. Examples of each dataset are respectively illustrated in Fig. \ref{Fig_illus_dataset_renoir} to Fig. \ref{Fig_illus_dataset_ours}. More detailed analysis of existing datasets can be found in \cite{DatasetXu}.
\subsection{Existing Dataset}
\subsubsection{RENOIR Dataset \cite{Anaya2014RENOIR}}
To the best of our knowledge, the RENOIR dataset is the first trial to construct real-world color image dataset with noisy and "ground-truth" image pairs. Three cameras are used to take photos of static indoor scenes with different ISO values. But the limitation of this dataset is that some image pairs exhibit misalignment and clear color differences due to the less refined post-processing step \cite{DatasetXu}.
\subsubsection{Nam Dataset \cite{Nam2016A}} Three different cameras are used to take photos of 11 static scenes. Each "ground truth" image is generated by capturing the same and unchanged scene for approximately 500 times and computing their mean value. The major problem with this dataset is that the images are mostly confined to printed scenes, whose statistical properties are similar \cite{DatasetXu}.
\subsubsection{DND Dataset \cite{Pl2017Benchmarking}} Four cameras are used to construct the DND dataset, but different from the previous two datasets, the authors utilize the Tobit regression to estimate the parameters of the noise process by accessing only two images. Careful post-processing step is conducted to correct the misalignment and remove residual low-frequency bias. The "ground-truth" images of this dataset are currently not available, but objective results of denoising methods can be obtained by submitting filtered images to the authors' website.
\subsubsection{Xu Dataset \cite{DatasetXu}} In order to address the limitations of above datasets, this new benchmark uses 5 different cameras and includes a broader variety of indoor scenes with more camera settings. Similar to \cite{Nam2016A}, each "ground-truth" image is also obtained by averaging the static images captured on the same scene. Besides, three volunteers are invited to manually remove outlier images that show clear misalignment or different illuminations.
\subsection{The Proposed Dataset}
\begin{figure}[htbp]
\graphicspath{{Illus_image/ours/}}
\centering
\subfigure[HUAWEI HONOR 6X]{
\label{Fig4}
\includegraphics[width=0.98in]{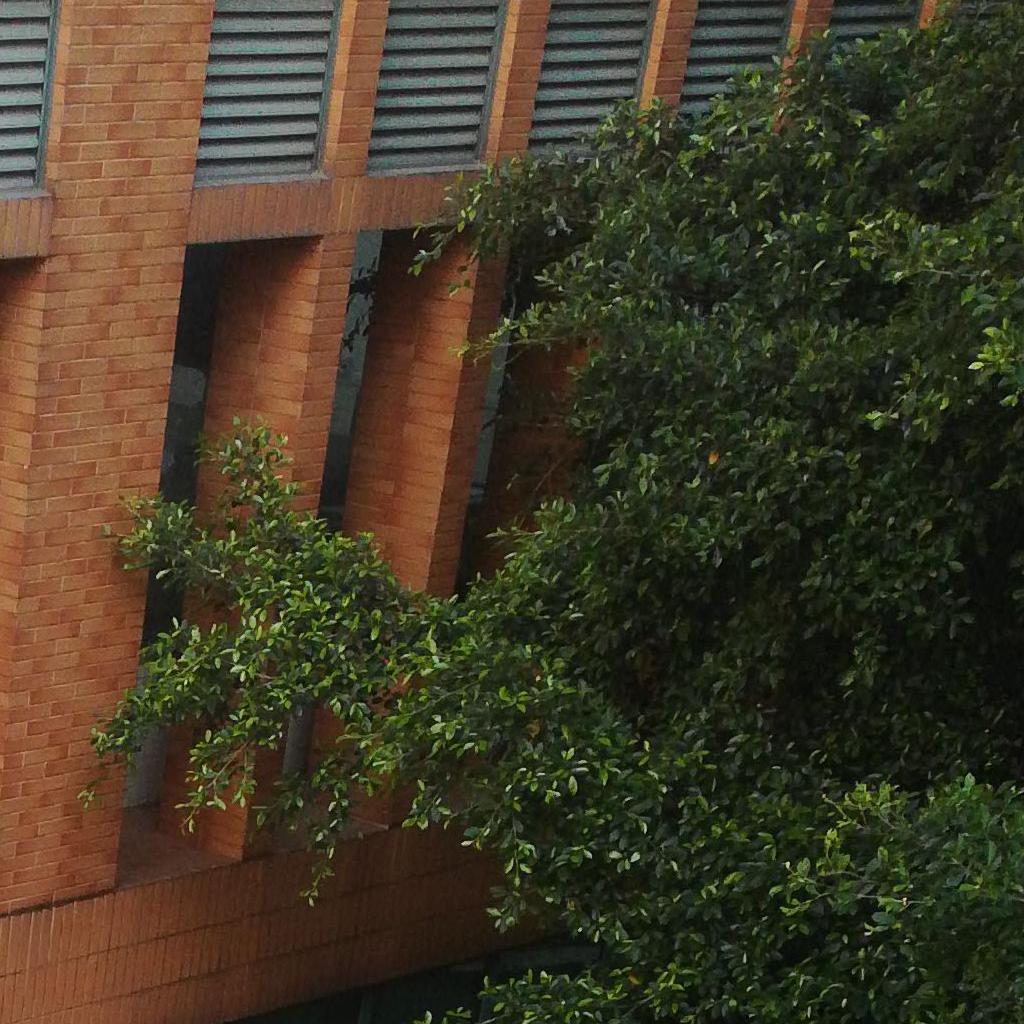}}
\subfigure[IPHONE 6S]{
\label{Fig4}
\includegraphics[width=0.98in]{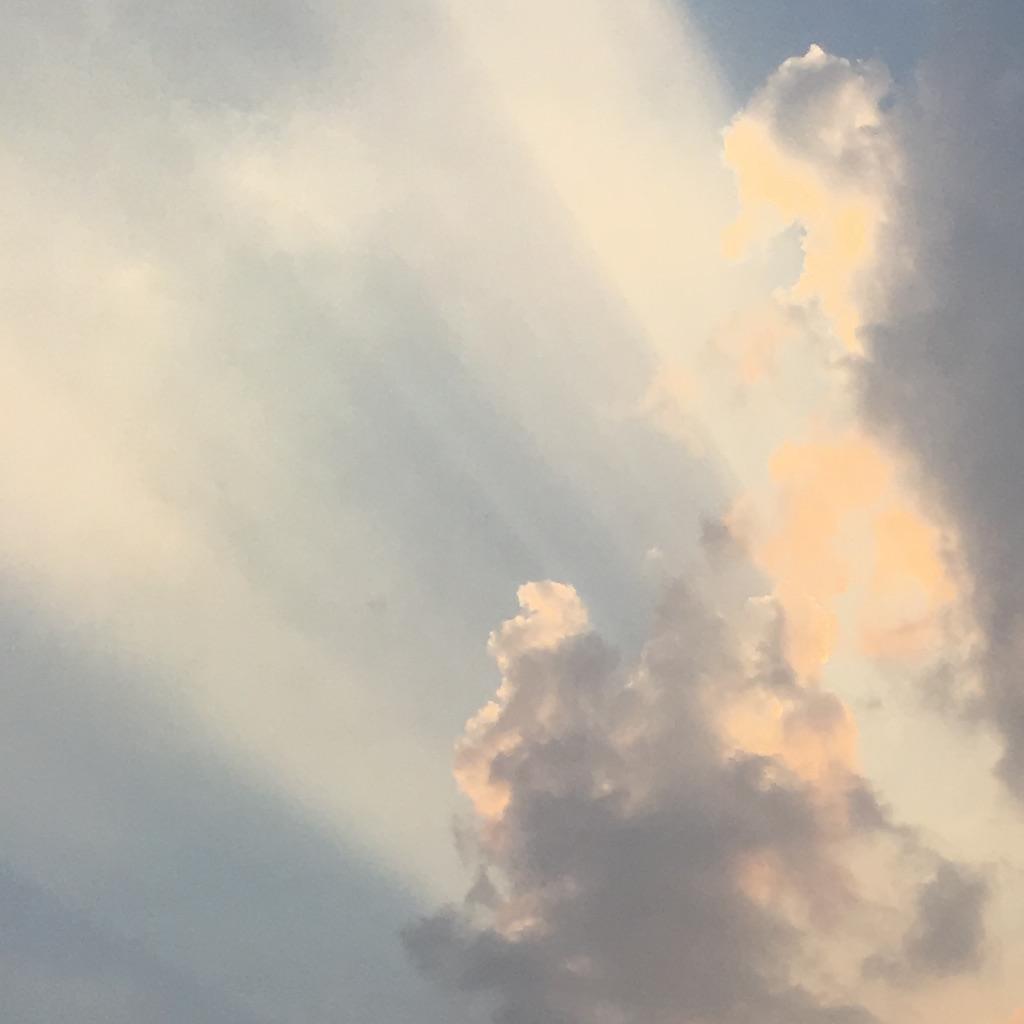}}
\subfigure[CANON 600D]{
\label{Fig5}
\includegraphics[width=0.98in]{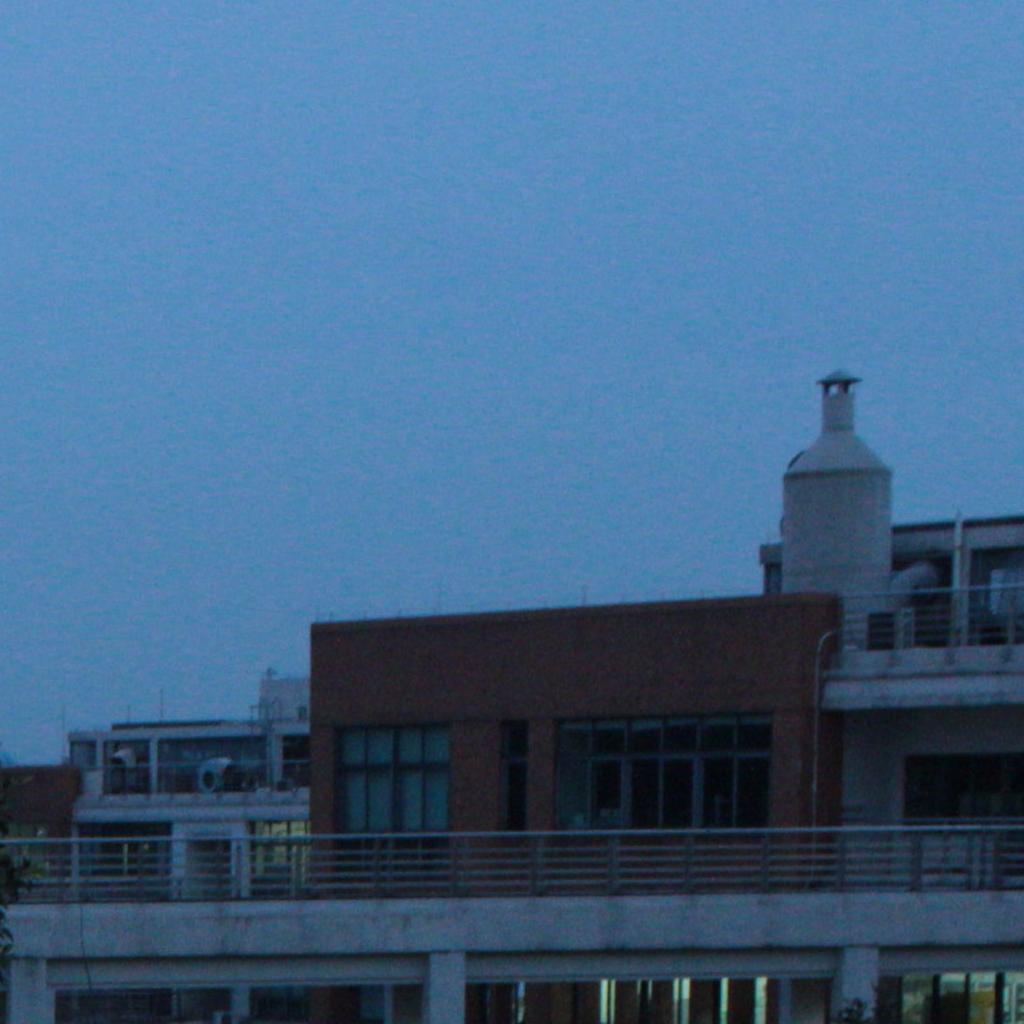}}

\caption{Some sample real noisy images of our newly constructed dataset.}

\label{Fig_illus_dataset_ours}
\end{figure}

The limitation of previously mentioned datasets is that they are based mainly on static indoor scenes where the lighting conditions can be manually controlled. But in real cases, many photos are captured on outdoor scenes where the objects and source of light are more complicated. In our dataset, six different cameras are used to take photos of both indoor and outdoor scenes. The same strategy of \cite{Nam2016A} and \cite{DatasetXu} is employed to generate "ground-truth" images due to its simplicity. But the images taken outdoor should be treated more carefully because natural objects such as flowers and trees are not completely motionless, and the variation of lighting condition can be intense due to the movement of cloud. Therefore, the shutter speed of cameras should be fast enough, and images that show very obvious misalignment and illumination differences are discarded. Each outdoor "ground-truth" image is obtained by averaging 30 to 60 images of the same scene. Several examples of the new dataset are illustrated in Fig. \ref{Fig_illus_dataset_ours}.
\begin{table}[htbp]
  \scriptsize
  \begin{threeparttable}
  \caption{Information of Five Real-World Color Image Datasets.}
  \centering
    \begin{tabular}{cccc}
    \toprule
    Dataset & Camera Brand & Number of Images & Sensor Size (mm) \\
    \midrule
    \multirow{3}[6]{*}{RENOIR} & CANON S90 & 40    & 7.4 $\times$ 5.6 \\
\cmidrule{2-4}          & CANON T3i & 40    & 22.3 $\times$ 14.9 \\
\cmidrule{2-4}          & XIAOMI Mi3 & 40    & 4.7 $\times$ 3.5 \\
    \midrule
    \multirow{3}[6]{*}{Nam} & CANON 5D MARK III & 3     & 36.0 $\times$ 24.0 \\
\cmidrule{2-4}          & NIKON D600 & 3     & 36.0 $\times$ 24.0 \\
\cmidrule{2-4}          & NIKON D800 & 9     & 36.0 $\times$ 24.0 \\
    \midrule
    \multirow{4}[8]{*}{DND} & SONY A7R & 13    & 36.0 $\times$ 24.0 \\
\cmidrule{2-4}          & OLYMPUS E-M10 & 13    & 17.3 $\times$ 13.0 \\
\cmidrule{2-4}          & SONY RX100 IV & 12    & 13.2 $\times$ 8.8 \\
\cmidrule{2-4}          & HUAWEI NEXUS 6P & 12    & 6.2 $\times$ 4.6 \\
    \midrule
    \multirow{5}[10]{*}{Xu} & CANON 5D & 10    & 36.0 $\times$ 24.0 \\
\cmidrule{2-4}          & CANON 80D & 6     & 22.5 $\times$ 15.0 \\
\cmidrule{2-4}          & CANON 600D & 5     & 22.3 $\times$ 14.9 \\
\cmidrule{2-4}          & NIKON D800 & 12    & 36.0 $\times$ 24.0 \\
\cmidrule{2-4}          & SONY A7II & 7     & 35.8 $\times$ 23.9 \\
    \midrule
    \multirow{6}[12]{*}{Ours} & HUAWEI HONOR 6X & 18    & 4.9 $\times$ 3.7 \\
\cmidrule{2-4}          & IPHONE 5S & 18    & 4.8 $\times$ 3.6 \\
\cmidrule{2-4}          & IPHONE 6S & 36    & 4.8 $\times$ 3.6 \\
\cmidrule{2-4}          & CANON 100D & 26    & 22.3 $\times$ 14.9 \\
\cmidrule{2-4}          & CANON 600D & 23    & 22.3 $\times$ 14.9 \\
\cmidrule{2-4}          & SONY A6500 & 18    & 15.6 $\times$ 23.5 \\
    \bottomrule
    \end{tabular}%
  \label{Table_dataset_description}%
  \end{threeparttable}
\end{table}%

\section{Experiments}

\subsection{Benchmark Datasets}
According to the description and analysis in Section III, the RENOIR and DND datasets are not used in our objective evaluations, because image pairs of the former exhibit some misalignment, while "ground-truth" images of the latter are not yet open access. But they will be included in our discussion on visual evaluations, since in real cases, "ground-truth" images are always not available. Besides, the image size of other three datasets are too large for some methods, thus four sub-datasets of cropped images are used, and a brief description of these sub-datasets is given in Table \ref{Table_info_subdataset}. All subdatasets are available at https://github.com/ZhaomingKong.
\begin{table}[htbp]
  \centering
  \scriptsize
  \caption{Information of four sub-datasets.}
    \begin{tabular}{cccc}
    \toprule
    Subdataset & Dataset & \# of Cropped Images & Image Size \\
    \midrule
    Dataset 1 & Nam \cite{Nam2016A}  & 15    & $512 \times 512$ \\
    \midrule
    Dataset 2 & Nam \cite{Nam2016A}  & 60    & $500 \times 500$ \\
    \midrule
    Dataset 3 & Xu  \cite{DatasetXu}  & 100   & $512 \times 512$ \\
    \midrule
    Dataset 4 & Ours \cite{Zhaoming} & 298   & $1024 \times 1024$ \\
    \bottomrule
    \end{tabular}%
  \label{Table_info_subdataset}%
\end{table}%

\subsection{Comparison Methods}
A comprehensive evaluation is conducted on several state-of-the-art methods, including MS-TSVD \cite{Zhaoming}, CBM3D \cite{Dabov2007Image}, 4DHOSVD1 \cite{rajwade2013image}, MCWNNM \cite{xu2017multi}, GID \cite{xu2018external}, TWSC \cite{TrilateralXu}, LLRT \cite{chang2017hyper}, LSCD \cite{tu2014collaborative}, DnCNN \cite{Harmeling2012Image} and TNRD \cite{Chen2015On}. The well-known commercial software Neat Image (NI) is included in our discussion on visual evaluation. In real cases, the parameters of all competing methods should be specified for each corrupted image, it is impractical because of the high computational complexity of some compared methods, but to better understand the effectiveness of CBM3D, we carefully tune the input noise level $\sigma$ of CBM3D for each image, and term this implementation 'CBM3D\_best' . In our experiments, code and implementation provided by the authors are used, and the average best results on each dataset are reported. All experiments are performed under MATLAB 2017a on a moderate laptop equipped with I5-8250U CPU of 1.8GHz and 8GB RAM.

\subsection{Objective Index}
There are several objective indexes \cite{wang2004image,zhang2011fsim} being applied to evaluate image quality by comparing filtered image and "ground-truth" image. In our experiments, two most commonly used indexes called Peak-to-Signal-Noise-Ratio (PSNR) and (Structural Similarity Index) SSIM \cite{wang2004image} are adopted. PSNR is a pixel-by-pixel comparison strategy that can be obtained by:
\begin{equation}\label{psnr}
\begin{split}
  & PSNR = 10\times log10(\frac{255^2}{mse}) \\
    &  mse  = \frac{||I_A - I_B||_F^2}{pixel \: number}
\end{split}
\end{equation}
where $I_A$ and $I_B$ are two compared images. Typically, the higher the PSNR and SSIM values, the better the results. But in our experiments, we will show that visual effects do not always correspond well with the objective index, partially because the human visual system (HSV) is not susceptible to the presence of noise.
\subsection{Experiment Results}
\subsubsection{Results on Dataset 1}
\begin{figure*}[htbp]
\graphicspath{{Illus_image/Compare_dataset_CC15/Combine/}}
\centering
\subfigure[Clean]{
\label{Fig4}
\includegraphics[width=1.1in]{Clean_demo1_combine_paint}}
\subfigure[Noisy]{
\label{Fig4}
\includegraphics[width=1.1in]{Noisy_demo1_combine}}
\subfigure[LSCD]{
\label{Fig4}
\includegraphics[width=1.1in]{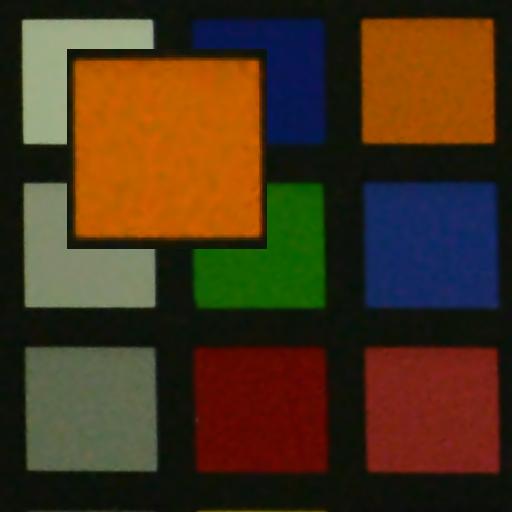}}
\subfigure[LLRT]{
\label{Fig5}
\includegraphics[width=1.1in]{LLRT_demo1_combine}}
\subfigure[GID]{
\label{Fig5}
\includegraphics[width=1.1in]{GID_demo1_combine}}\\

\subfigure[TWSC]{
\label{Fig4}
\includegraphics[width=1.1in]{TWSC_demo1_combine}}
\subfigure[MCWNNM]{
\label{Fig4}
\includegraphics[width=1.1in]{MCWNNM_demo1_combine}}
\subfigure[4DHOSVD1]{
\label{Fig4}
\includegraphics[width=1.1in]{4DHOSVD1_demo1_combine}}
\subfigure[CBM3D]{
\label{Fig4}
\includegraphics[width=1.1in]{CBM3D_demo1_combine}}
\subfigure[MS-TSVD]{
\label{Fig4}
\includegraphics[width=1.1in]{MS-TSVD_demo1_combine}}

\caption{Denoising images of compared methods on Dataset 1. The camera is CANON D600 with ISO = 3200. Please zoom-in for better view.}

\label{Fig_compare_dataset1_demo1}
\end{figure*}

\begin{figure*}[htbp]
\graphicspath{{Illus_image/Compare_dataset_CC15/Combine/}}
\centering
\subfigure[Clean]{
\label{Fig4}
\includegraphics[width=1.1in]{Clean_demo2_combine_paint}}
\subfigure[Noisy]{
\label{Fig4}
\includegraphics[width=1.1in]{Noisy_demo2_combine}}
\subfigure[LSCD]{
\label{Fig4}
\includegraphics[width=1.1in]{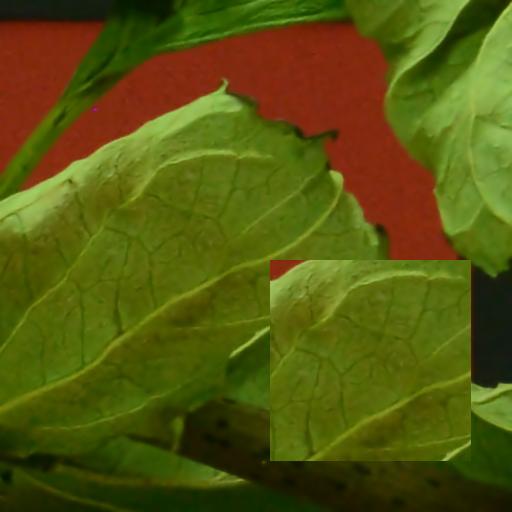}}
\subfigure[LLRT]{
\label{Fig5}
\includegraphics[width=1.1in]{LLRT_demo2_combine}}
\subfigure[GID]{
\label{Fig5}
\includegraphics[width=1.1in]{GID_demo2_combine}}\\

\subfigure[TWSC]{
\label{Fig4}
\includegraphics[width=1.1in]{TWSC_demo2_combine}}
\subfigure[MCWNNM]{
\label{Fig4}
\includegraphics[width=1.1in]{MCWNNM_demo2_combine}}
\subfigure[4DHOSVD1]{
\label{Fig4}
\includegraphics[width=1.1in]{4DHOSVD1_demo2_combine}}
\subfigure[CBM3D]{
\label{Fig4}
\includegraphics[width=1.1in]{CBM3D_demo2_combine}}
\subfigure[MS-TSVD]{
\label{Fig4}
\includegraphics[width=1.1in]{MS-TSVD_demo2_combine}}

\caption{Denoising images of compared methods on Dataset 1. The camera is NIKON D800 with ISO = 1600. Please zoom-in to for better view.}

\label{Fig_compare_dataset1_demo2}
\end{figure*}

To show the robustness of all compared methods, the PSNR value of each image is listed in Table \ref{Table_dataset1}. The average computational time is also provided. Apart from CBM3D, LSCD and TNRD that use C++ mex-function with parallelization technique, other compared methods are purely implemented in MATLAB. For MS-TSVD, the authors also provide a C++ implementation\footnote{https://github.com/ZhaomingKong/color\_image\_denoising} that can reduce its computational time to less than 8 seconds. From Table \ref{Table_dataset1}, one can see that the simple MS-TSVD method is able to produce very competitive performance in terms of both effectiveness and efficiency. Fig. \ref{Fig_compare_dataset1_demo1} and Fig. \ref{Fig_compare_dataset1_demo2} show the denoised images captured by CANON D600 at ISO = 3200 and NIKON D800 at ISO = 1600, respectively. The visual evaluation shows that the representative low-rank based method LLRT and the sparse coding scheme TWSC produce better results in homogenous regions, because the underlying clean patches share similar feature, and thus can be approximated by a low-rank or sparse coding problem. But as is illustrated in Fig. \ref{Fig_compare_dataset1_demo2}, when the ground truth image contains more details, it may be risky to employ the low-rank approximation strategy, and the clear over-smooth result contradicts the improvement in PSNR value. Compared with CBM3D and MS-TSVD that utilize global patch representation, the local 4DHOSVD transform is more easily affected by the presence of noise, which is incorporated in the training process of color mode transform.
\begin{table*}[htbp]
  \centering
  \ssmall
  \caption{PSNR (dB) results and SPEED comparison of different methods on Dataset 1. Three best results are bolded.}
    \begin{tabular}{cccccccccccc}
    \toprule
    Camara & LSCD  & LLRT  & TNRD  & DnCNN & GID   & TWSC  & MCWNNM & CBM3D & CBM3D\_best & 4DHOSVD1 & MS-TSVD \\
    \midrule
    \multirow{2}[4]{*}{CANON 5D } & 37.86  & 39.23  & 39.51  & 37.26  & 40.82  & 40.55  & \textbf{41.20 } & 40.77  & \textbf{40.96 } & 40.22  & \textbf{40.79 } \\
\cmidrule{2-12}          & 36.21  & 36.31  & 36.47  & 34.13  & 37.19  & 35.92  & 37.25  & \textbf{37.31 } & \textbf{37.31 } & 36.97  & \textbf{37.37 } \\
\cmidrule{2-12}    ISO = 3200 & 35.52  & 35.93  & 36.45  & 34.09  & 36.92  & 35.15  & 36.48  & \textbf{36.98 } & \textbf{37.15 } & 36.55  & \textbf{37.01 } \\
    \midrule
    \multirow{2}[4]{*}{NIKON D600 } & 34.65  & 34.74  & 34.79  & 33.62  & 35.32  & \textbf{35.36 } & \textbf{35.54 } & 35.21  & \textbf{35.38 } & 35.02  & 35.29  \\
\cmidrule{2-12}          & 36.26  & 36.83  & 36.37  & 34.48  & 36.62  & \textbf{37.09 } & \textbf{37.03 } & 36.76  & 36.81  & 36.60  & \textbf{36.95 } \\
\cmidrule{2-12}    ISO = 3200 & 38.24  & 40.58  & 39.49  & 35.41  & 38.68  & \textbf{41.13 } & 39.56  & 40.13  & \textbf{40.45 } & 39.78  & \textbf{40.93 } \\
    \midrule
    \multirow{2}[4]{*}{NIKON D800 } & 37.90  & 37.39  & 38.11  & 35.79  & 38.88  & \textbf{39.36 } & \textbf{39.26 } & 39.02  & \textbf{39.25 } & 38.85  & 39.21  \\
\cmidrule{2-12}          & 38.88  & 40.27  & 40.52  & 36.08  & 40.66  & \textbf{41.91 } & 41.45  & \textbf{41.65 } & \textbf{41.65 } & 41.35  & \textbf{41.98 } \\
\cmidrule{2-12}    ISO = 1600 & 38.32  & 37.78  & 38.17  & 35.48  & 39.20  & 38.81  & \textbf{39.54 } & 39.40  & \textbf{39.59 } & 39.11  & \textbf{39.54 } \\
    \midrule
    \multirow{2}[4]{*}{NIKON D800 } & 37.45  & 39.79  & 37.69  & 34.08  & 37.92  & \textbf{40.27 } & 38.94  & 39.59  & \textbf{39.86 } & 39.24  & \textbf{39.98 } \\
\cmidrule{2-12}          & 36.49  & 37.34  & 35.90  & 33.70  & 36.62  & 37.22  & 37.40  & \textbf{37.49 } & \textbf{37.54 } & 37.28  & \textbf{37.65 } \\
\cmidrule{2-12}    ISO = 3200 & 37.73  & \textbf{41.03 } & 38.21  & 33.31  & 37.64  & \textbf{42.09 } & 39.42  & 39.47  & \textbf{40.38 } & 39.47  & 40.05  \\
    \midrule
    \multirow{2}[4]{*}{NIKON D800 } & 32.33  & \textbf{35.09 } & 32.81  & 29.83  & 33.01  & \textbf{35.53 } & \textbf{34.85 } & 34.13  & \textbf{34.85 } & 34.40  & 34.50  \\
\cmidrule{2-12}          & 32.55  & \textbf{34.05 } & 32.33  & 30.55  & 32.93  & \textbf{34.15 } & \textbf{33.97 } & 33.73  & 33.92  & 33.81  & 33.93  \\
\cmidrule{2-12}    ISO = 6400 & 32.62  & \textbf{34.11 } & 32.29  & 30.09  & 32.96  & 33.93  & 33.97  & 33.85  & \textbf{34.16 } & \textbf{34.01 } & \textbf{34.01 } \\
    \midrule
    Average & 36.20  & 37.36  & 36.61  & 33.86  & 37.02  & \textbf{37.90 } & 37.72  & 37.70  & \textbf{37.95 } & 37.51  & \textbf{37.95 } \\
    \midrule
    Time (s) & 9.8   & 1168.9 & 5.6   & 98.6  & 55.6  & 498.8 & 298.8 & 6.8   & 6.8   & 130.8 & 98.8 \\
    \bottomrule
    \end{tabular}%
  \label{Table_dataset1}%
\end{table*}%

\subsubsection{Results on Dataset 2 and Dataset 3}
\begin{table*}[htbp]
  \centering
  \caption{Average PSNR (dB) and SSIM if different denoising methods on Dataset 2 and Dataset 3. The best results are bolded.}
    \begin{tabular}{cccccccccc}
    \toprule
    Dataset & Index & LLRT  & GID   & TWSC  & MCWNNM & 4DHOSVD1 & CBM3D & CBM3D\_best & MS-TSVD \\
    \midrule
    \multirow{2}[4]{*}{Dataset 2} & PSNR  & 38.51  & 38.41  & 39.66  & 39.03  & 39.15  & 39.40  & 39.68  & \textbf{39.75} \\
\cmidrule{2-10}          & SSIM  & 0.9636  & 0.9633  & 0.9759  & 0.9698  & 0.9729  & 0.9740  & \textbf{0.9775}  & 0.9756 \\
    \midrule
    \multirow{2}[4]{*}{Dataset 3} & PSNR  & 38.51 & 38.37 & 38.62 & 38.51 & 38.51 & 38.69 & \textbf{38.81} & \textbf{38.82} \\
\cmidrule{2-10}          & SSIM  & 0.9707 & 0.9675 & 0.9674 & 0.9671 & 0.9673 & 0.9694 & \textbf{0.9712} & 0.9694 \\
    \bottomrule
    \end{tabular}%
  \label{Table_dataset2_and_3}%
\end{table*}%

\begin{figure*}[htbp]
\graphicspath{{Illus_image/Compare_dataset_CC60/Combine/}}
\centering
\subfigure[Clean]{
\label{Fig4}
\includegraphics[width=1.1in]{Clean_demo1_combine_paint}}
\subfigure[Noisy]{
\label{Fig4}
\includegraphics[width=1.1in]{Noisy_demo1_combine}}
\subfigure[NI]{
\label{Fig4}
\includegraphics[width=1.1in]{NI_demo1_combine}}
\subfigure[LLRT]{
\label{Fig5}
\includegraphics[width=1.1in]{LLRT_demo1_combine}}
\subfigure[GID]{
\label{Fig5}
\includegraphics[width=1.1in]{GID_demo1_combine}}\\

\subfigure[TWSC]{
\label{Fig4}
\includegraphics[width=1.1in]{TWSC_demo1_combine}}
\subfigure[MCWNNM]{
\label{Fig4}
\includegraphics[width=1.1in]{MCWNNM_demo1_combine}}
\subfigure[4DHOSVD1]{
\label{Fig4}
\includegraphics[width=1.1in]{4DHOSVD1_demo1_combine}}
\subfigure[CBM3D]{
\label{Fig4}
\includegraphics[width=1.1in]{CBM3D_demo1_combine}}
\subfigure[MS-TSVD]{
\label{Fig4}
\includegraphics[width=1.1in]{MS-TSVD_demo1_combine}}

\caption{Denoising images of compared methods on Dataset 2. The camera is NIKON D800 with ISO = 1600. Please zoom-in for better view.}

\label{Fig_compare_dataset_CC60}
\end{figure*}

\begin{figure*}[htbp]
\graphicspath{{Illus_image/Compare_dataset_Xu/Combine/}}
\centering
\subfigure[Clean]{
\label{Fig4}
\includegraphics[width=1.1in]{Clean_demo1_combine_paint}}
\subfigure[Noisy]{
\label{Fig4}
\includegraphics[width=1.1in]{Noisy_demo1_combine}}
\subfigure[NI]{
\label{Fig4}
\includegraphics[width=1.1in]{NI_demo1_combine}}
\subfigure[LLRT]{
\label{Fig5}
\includegraphics[width=1.1in]{LLRT_demo1_combine}}
\subfigure[GID]{
\label{Fig5}
\includegraphics[width=1.1in]{GID_demo1_combine}}\\

\subfigure[TWSC]{
\label{Fig4}
\includegraphics[width=1.1in]{TWSC_demo1_combine}}
\subfigure[MCWNNM]{
\label{Fig4}
\includegraphics[width=1.1in]{MCWNNM_demo1_combine}}
\subfigure[4DHOSVD1]{
\label{Fig4}
\includegraphics[width=1.1in]{4DHOSVD1_demo1_combine}}
\subfigure[CBM3D]{
\label{Fig4}
\includegraphics[width=1.1in]{CBM3D_demo1_combine}}
\subfigure[MS-TSVD]{
\label{Fig4}
\includegraphics[width=1.1in]{MS-TSVD_demo1_combine}}

\caption{Denoising images of compared methods on Dataset 3. The camera is CANON 5D with ISO = 6400. Please zoom-in for better view.}

\label{Fig_compare_dataset_Xu}
\end{figure*}

In Table \ref{Table_dataset2_and_3}, we list the results of several most competitive methods on Dataset 2 and Dataset 3. It can be seen that on these two datasets, MS-TSVD also achieves the most competitive performance, it is almost the upper bound of the effectiveness of CBM3D. To further demonstrate the observation made in our experiments of the Dataset 1, two images that contain rich details and large homogenous regions are chosen respectively from Dataset 2 and Dataset 3. Visual evaluations are given in Fig. \ref{Fig_compare_dataset_CC60} and Fig. \ref{Fig_compare_dataset_Xu}.

\subsubsection{Results on Dataset 4}
\begin{table*}[htbp]
  \centering
  \ssmall
  \caption{Average PSNR (dB) and SSIM if different denoising methods on our newly constructed dataset. The best results are bolded.}
    \begin{tabular}{ccccccccccc}
    \toprule
    Camera & Cropped Images & Index & LLRT  & GID   & TWSC  & MCWNNM & 4DHOSVD1 & CBM3D & CBM3D\_best & MS-TSVD \\
    \midrule
    \multirow{2}[4]{*}{HUAWEI HONOR 6X} & \multirow{2}[4]{*}{30} & PSNR  & 39.54 & 39.52  & 39.71  & 39.46 & 39.82 & 39.97 & \textbf{40.48} & 40.08 \\
\cmidrule{3-11}          &       & SSIM  & 0.9669 & 0.9653 & 0.9651 & 0.9610  & 0.9658  & 0.9669  & \textbf{0.9740 } & 0.9674  \\
    \midrule
    \multirow{2}[4]{*}{IPHONE 5S} & \multirow{2}[4]{*}{36} & PSNR  & 40.02 & 40.12  & 40.27  & 39.87 & 40.68 & 40.77 & \textbf{41.25} & 40.84 \\
\cmidrule{3-11}          &       & SSIM  & 0.9676 & 0.9642 & 0.9617 & 0.9567  & 0.9664  & 0.9668  & \textbf{0.9758 } & 0.9668  \\
    \midrule
    \multirow{2}[4]{*}{IPHONE 6S} & \multirow{2}[4]{*}{67} & PSNR  & 39.72 & 40.16  & 40.12  & 40.18 & 40.36 & 40.55 & \textbf{41.16} & 40.53 \\
\cmidrule{3-11}          &       & SSIM  & 0.9663 & 0.9670  & 0.9619  & 0.9628  & 0.9671  & 0.9693  & \textbf{0.9783 } & 0.9674  \\
    \midrule
    \multirow{2}[4]{*}{CANON 100D} & \multirow{2}[4]{*}{55} & PSNR  & 41.84 & 40.86 & 41.65 & 41.47 & 41.41 & 41.69 & \textbf{42.08} & 41.99 \\
\cmidrule{3-11}          &       & SSIM  & 0.9784 & 0.9743  & 0.9767  & 0.9774  & 0.9771  & 0.9780  & \textbf{0.9808 } & 0.9794  \\
    \midrule
    \multirow{2}[4]{*}{CANON 600D} & \multirow{2}[4]{*}{25} & PSNR  & 42.53 & 41.60  & 42.52  & 42.07 & 42.14 & 42.54 & \textbf{42.89} & 42.75 \\
\cmidrule{3-11}          &       & SSIM  & 0.9816 & 0.9790  & 0.9824  & 0.9795  & 0.9810  & 0.9836  & \textbf{0.9851 } & 0.9840  \\
    \midrule
    \multirow{2}[4]{*}{SONY A6500} & \multirow{2}[4]{*}{36} & PSNR  & 45.71 & 44.94  & 45.48  & 45.37 & 45.56 & 45.7  & 45.81 & \textbf{45.89} \\
\cmidrule{3-11}          &       & SSIM  & 0.9899 & 0.9887  & 0.9896  & 0.9894  & 0.9901  & 0.9902  & \textbf{0.9904 } & 0.9903  \\
    \bottomrule
    \end{tabular}%
  \label{Table_dataset4}%
\end{table*}%

\begin{figure*}[htbp]
\graphicspath{{Illus_image/Compare_dataset_ours/Combine/}}
\centering
\subfigure[Clean]{
\label{Fig4}
\includegraphics[width=1.1in]{Clean_demo1_combine_paint}}
\subfigure[Noisy]{
\label{Fig4}
\includegraphics[width=1.1in]{Noisy_demo1_combine}}
\subfigure[LLRT]{
\label{Fig4}
\includegraphics[width=1.1in]{LLRT_demo1_combine}}
\subfigure[NI]{
\label{Fig4}
\includegraphics[width=1.1in]{NI_demo1_combine}}
\subfigure[GID]{
\label{Fig5}
\includegraphics[width=1.1in]{GID_demo1_combine}}\\
\subfigure[TWSC]{
\label{Fig5}
\includegraphics[width=1.1in]{TWSC_demo1_combine}}
\subfigure[MCWNNM]{
\label{Fig4}
\includegraphics[width=1.1in]{MCWNNM_demo1_combine}}
\subfigure[4DHOSVD1]{
\label{Fig4}
\includegraphics[width=1.1in]{4DHOSVD1_demo1_combine}}
\subfigure[CBM3D]{
\label{Fig4}
\includegraphics[width=1.1in]{CBM3D_demo1_combine}}
\subfigure[MS-TSVD]{
\label{Fig4}
\includegraphics[width=1.1in]{MS-TSVD_demo1_combine}}

\caption{Denoising images of compared methods on Dataset 3. The camera is HUAWEI HONOR 6X with auto mode. Please zoom-in for better view.}

\label{Fig_compare_dataset_ours_demo1}
\end{figure*}

\begin{figure*}[htbp]
\graphicspath{{Illus_image/Compare_dataset_ours/Combine/}}
\centering
\subfigure[Clean]{
\label{Fig4}
\includegraphics[width=1.1in]{Clean_demo2_combine_paint}}
\subfigure[Noisy]{
\label{Fig4}
\includegraphics[width=1.1in]{Noisy_demo2_combine}}
\subfigure[LLRT]{
\label{Fig4}
\includegraphics[width=1.1in]{LLRT_demo2_combine}}
\subfigure[NI]{
\label{Fig4}
\includegraphics[width=1.1in]{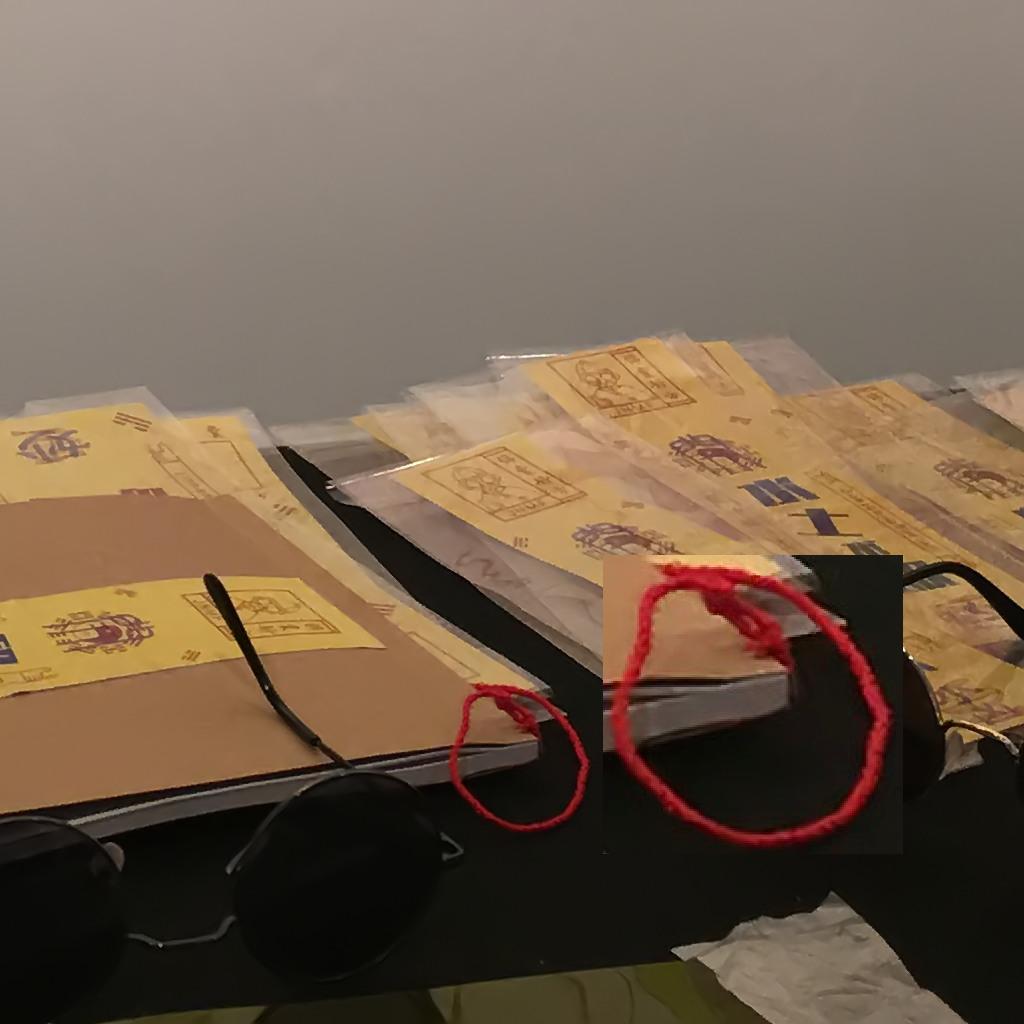}}
\subfigure[GID]{
\label{Fig5}
\includegraphics[width=1.1in]{GID_demo2_combine}}\\
\subfigure[TWSC]{
\label{Fig5}
\includegraphics[width=1.1in]{TWSC_demo2_combine}}
\subfigure[MCWNNM]{
\label{Fig4}
\includegraphics[width=1.1in]{MCWNNM_demo2_combine}}
\subfigure[4DHOSVD1]{
\label{Fig4}
\includegraphics[width=1.1in]{4DHOSVD1_demo2_combine}}
\subfigure[CBM3D]{
\label{Fig4}
\includegraphics[width=1.1in]{CBM3D_demo2_combine}}
\subfigure[MS-TSVD]{
\label{Fig4}
\includegraphics[width=1.1in]{MS-TSVD_demo2_combine}}

\caption{Denoising images of compared methods on Dataset 3. The camera is IPHONE 6S with auto mode. Please zoom-in for better view.}

\label{Fig_compare_dataset_ours_demo2}
\end{figure*}

\begin{figure*}[htbp]
\graphicspath{{Illus_image/Compare_dataset_ours/Combine/}}
\centering
\subfigure[Clean]{
\label{Fig4}
\includegraphics[width=1.1in]{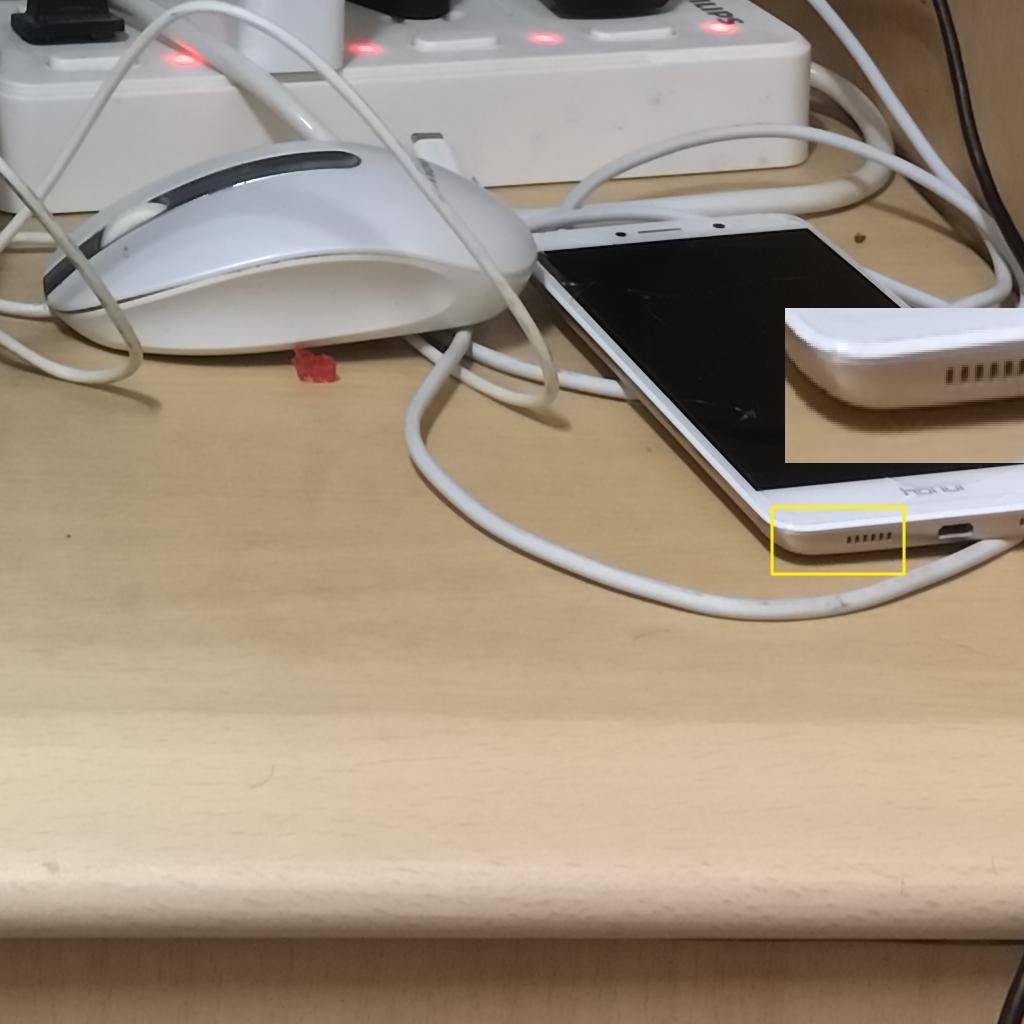}}
\subfigure[Noisy]{
\label{Fig4}
\includegraphics[width=1.1in]{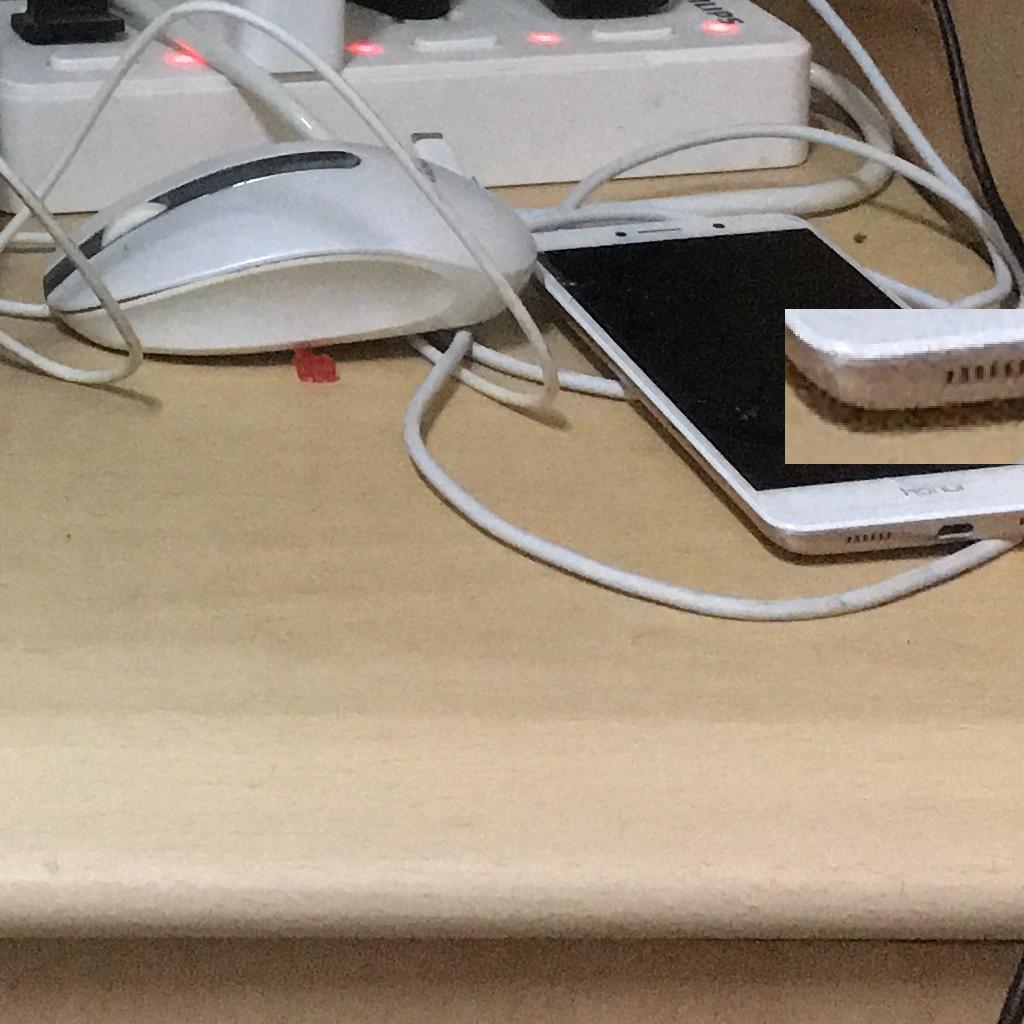}}
\subfigure[LLRT]{
\label{Fig4}
\includegraphics[width=1.1in]{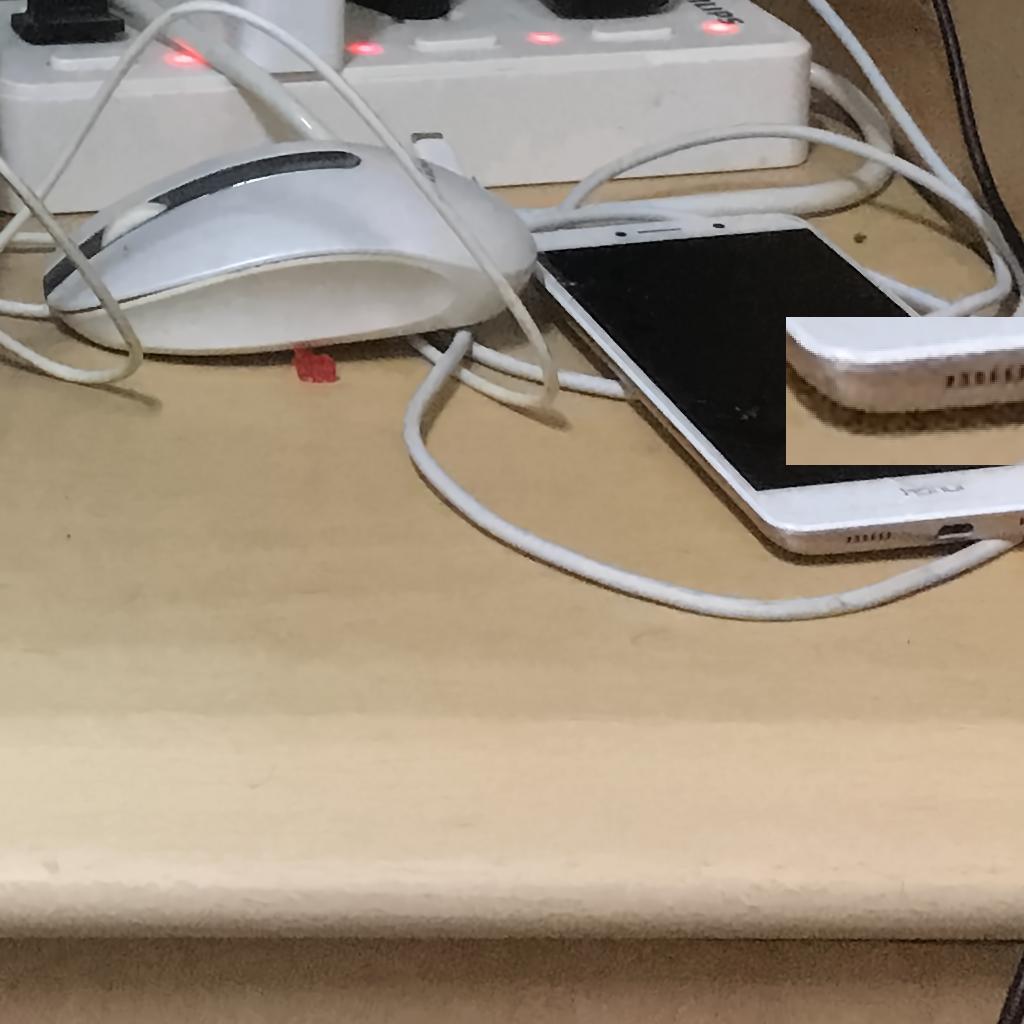}}
\subfigure[NI]{
\label{Fig4}
\includegraphics[width=1.1in]{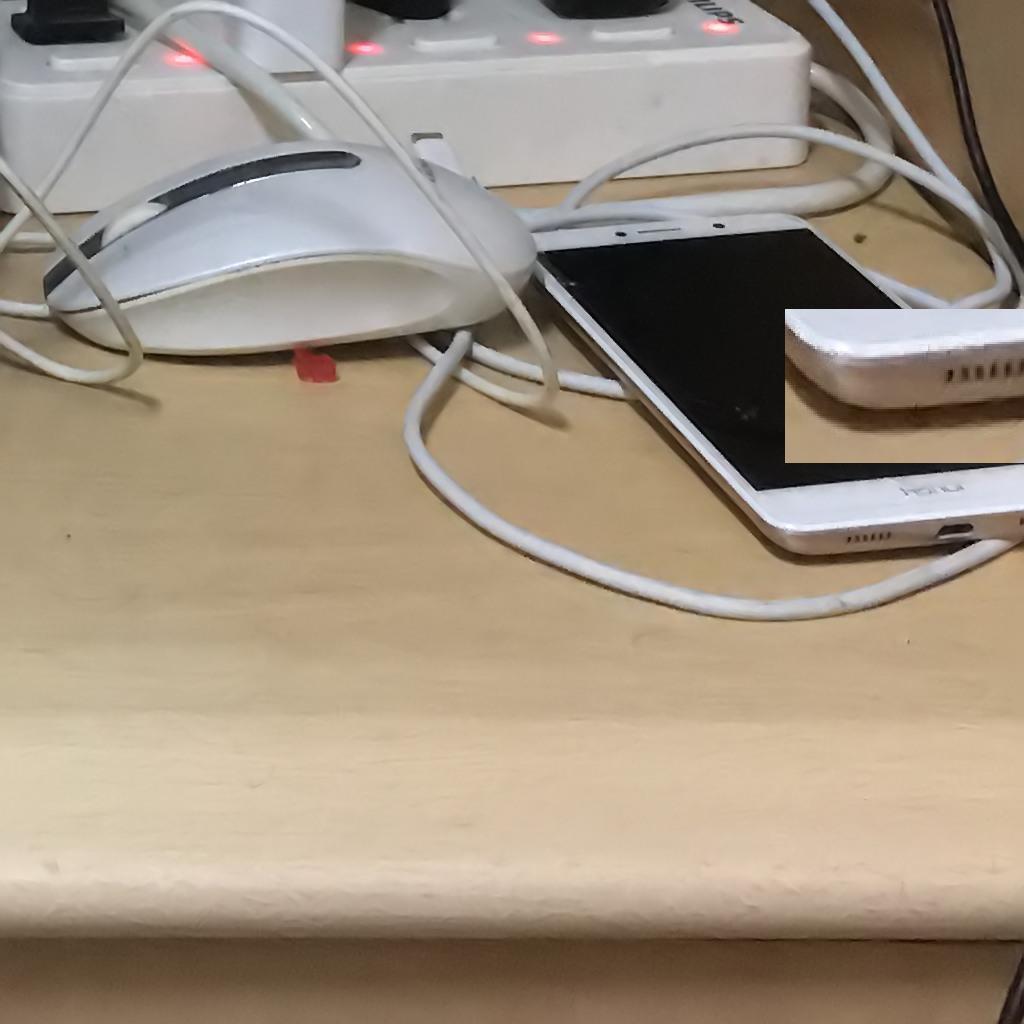}}
\subfigure[GID]{
\label{Fig5}
\includegraphics[width=1.1in]{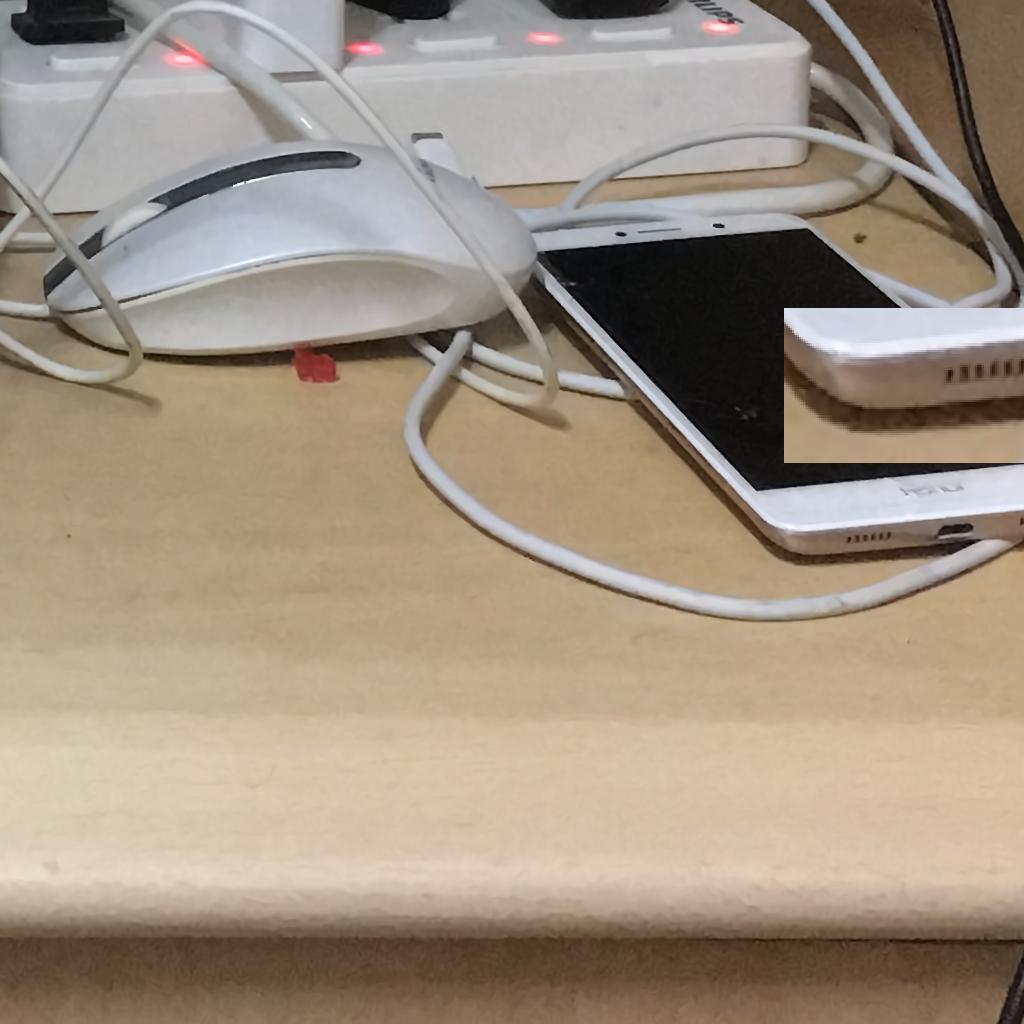}}\\
\subfigure[TWSC]{
\label{Fig5}
\includegraphics[width=1.1in]{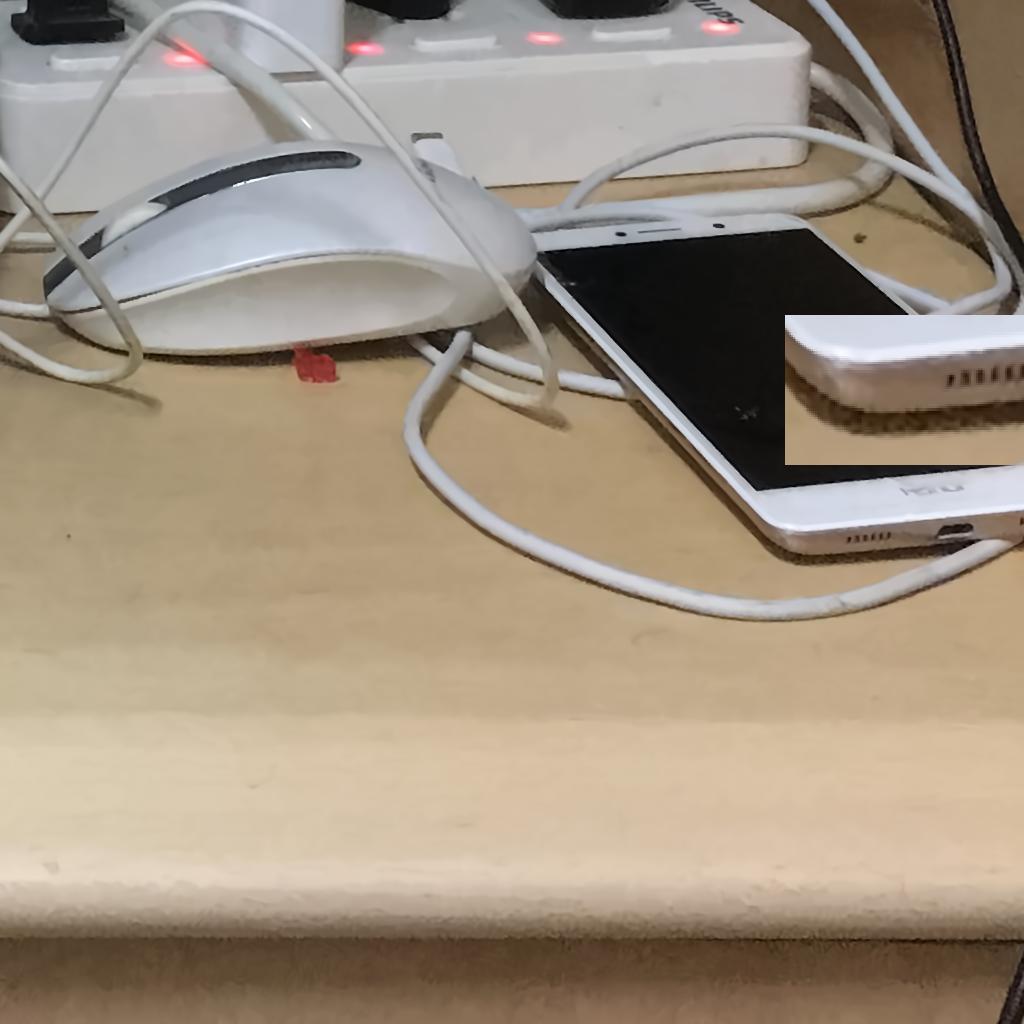}}
\subfigure[MCWNNM]{
\label{Fig4}
\includegraphics[width=1.1in]{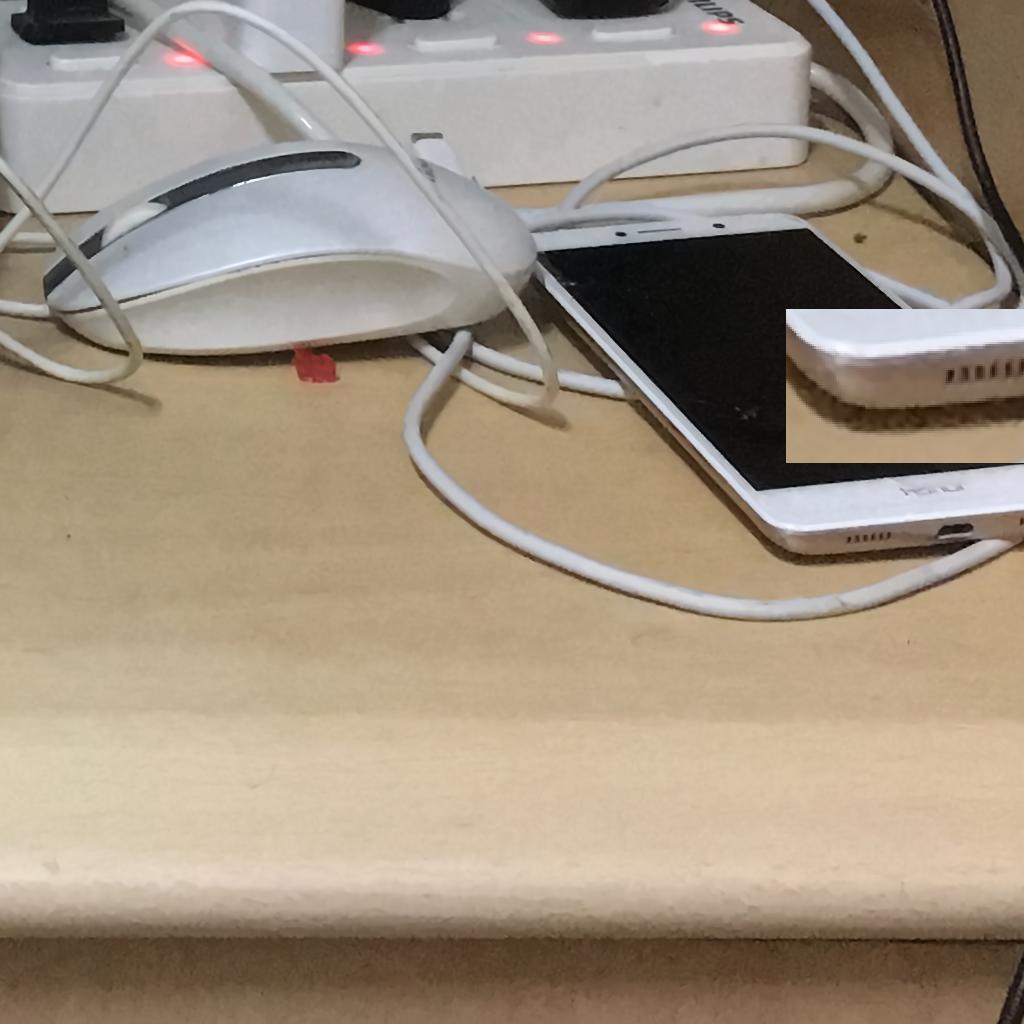}}
\subfigure[4DHOSVD1]{
\label{Fig4}
\includegraphics[width=1.1in]{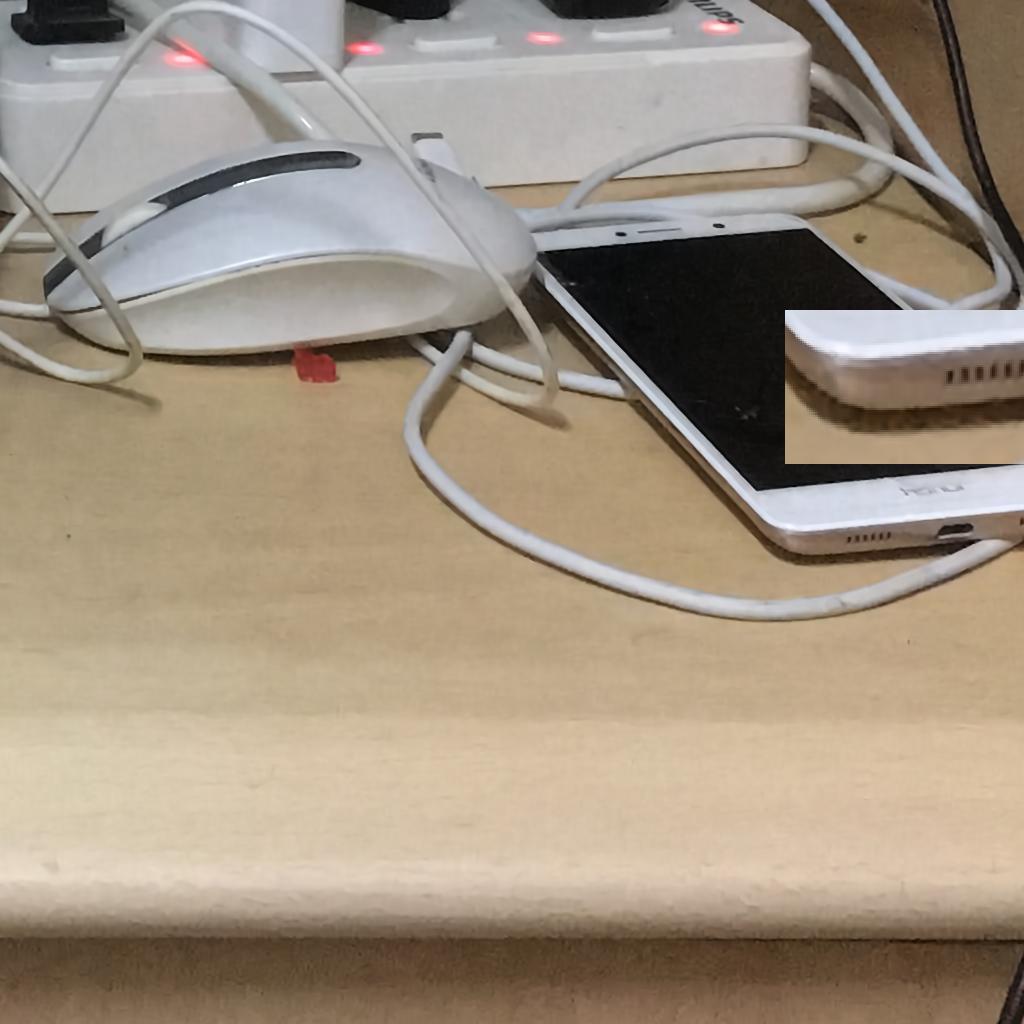}}
\subfigure[CBM3D]{
\label{Fig4}
\includegraphics[width=1.1in]{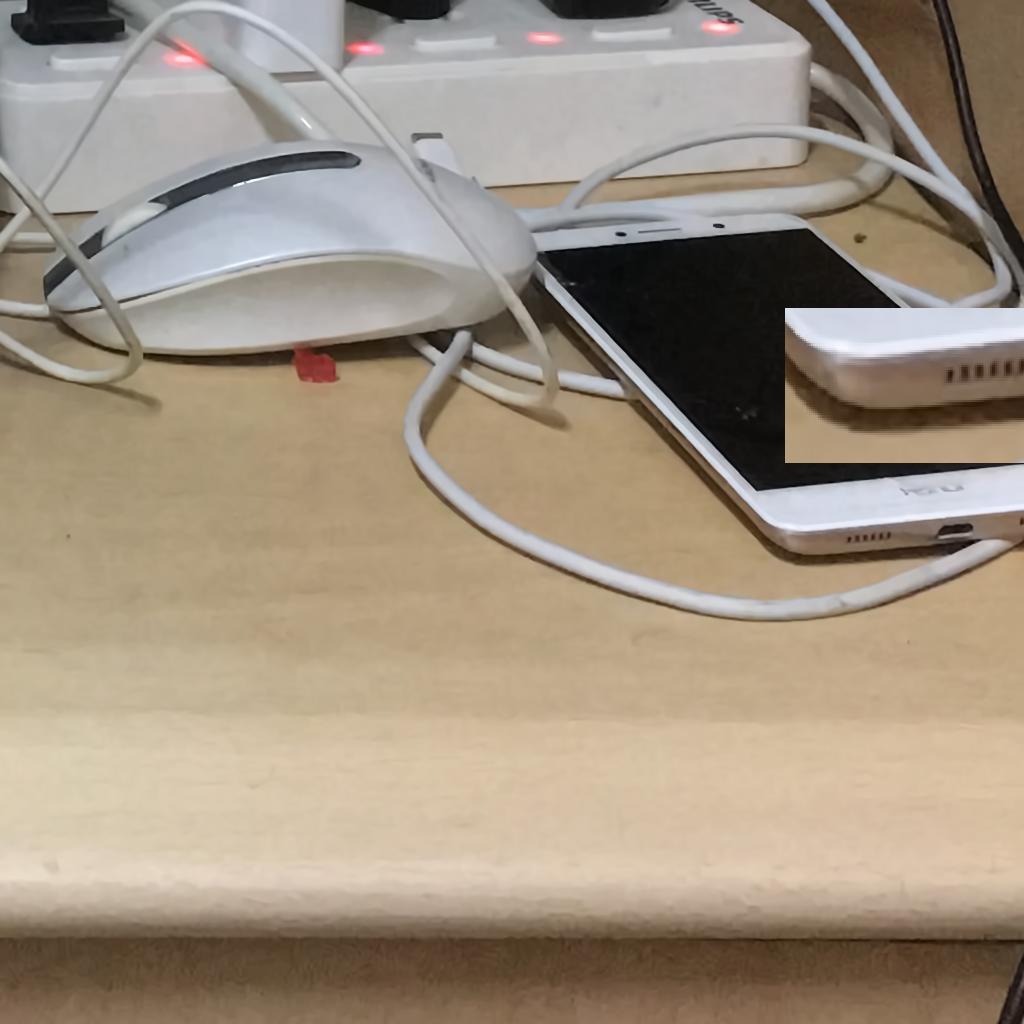}}
\subfigure[MS-TSVD]{
\label{Fig4}
\includegraphics[width=1.1in]{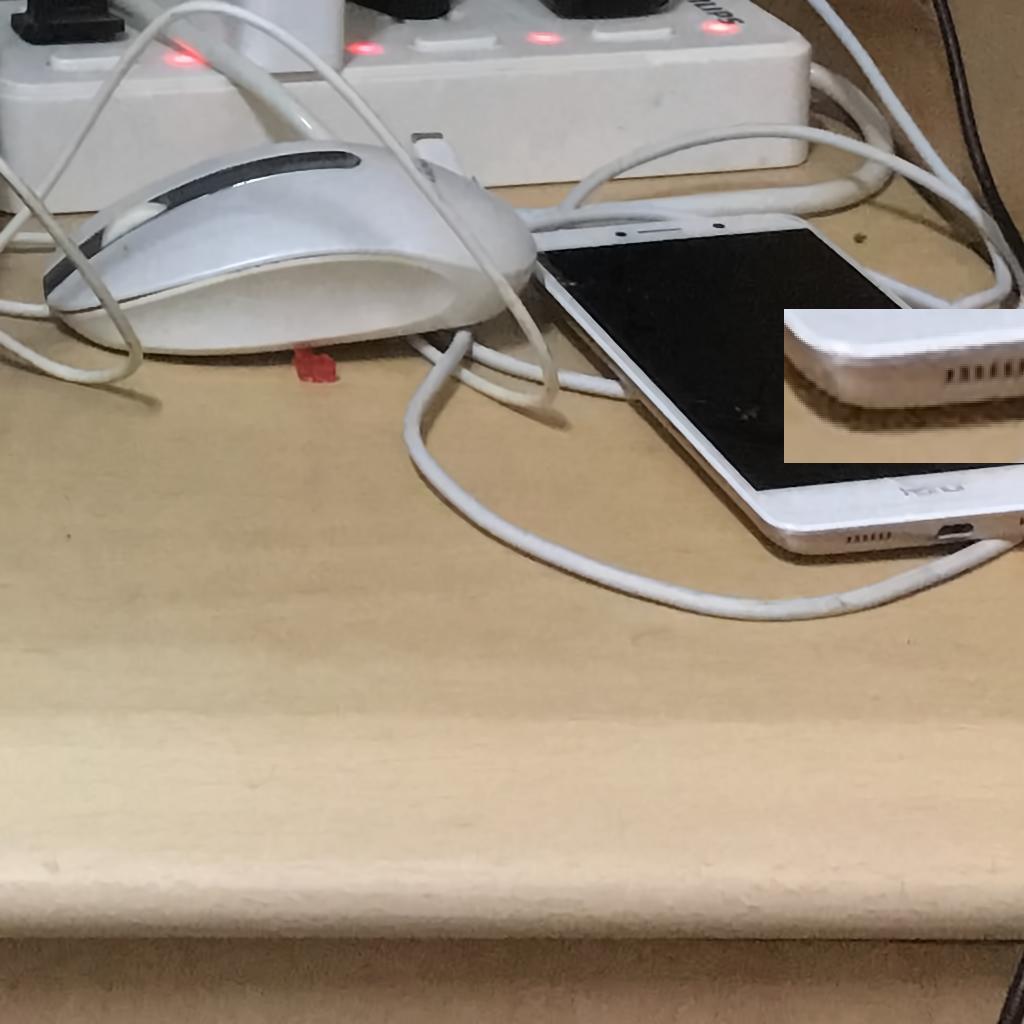}}

\caption{Denoising images of compared methods on Dataset 3. The camera is IPHONE 6S with auto mode. Please zoom-in for better view.}

\label{Fig_compare_dataset_ours_demo3}
\end{figure*}

Cropped images of this dataset is four times as large as that of the above datasets, thus several efficient and competitive methods are chosen according to our previous experiments. Average PSNR and SSIM values are listed in Table \ref{Table_dataset4}, while visual evaluations are given in Fig. \ref{Fig_compare_dataset_ours_demo1}, Fig. \ref{Fig_compare_dataset_ours_demo2} and Fig. \ref{Fig_compare_dataset_ours_demo3}. Comparing Table \ref{Table_dataset4} with Table \ref{Table_dataset1} and Table \ref{Table_dataset2_and_3}, it can be seen that no competing methods can consistently outperform the state-of-the-art CBM3D, especially when the image size is large and contains more natural outdoor scenes. Fig. \ref{Fig_compare_dataset_ours_demo2} shows the slight over-smooth effects of CBM3D, largely because its pre-defined transform is less adaptive, but this drawback is offset by its robustness to local variation, as is illustrated in Fig. \ref{Fig_compare_dataset_ours_demo3}. Interestingly, we observe in both Fig. \ref{Fig_compare_dataset_ours_demo1} and Fig. \ref{Fig_compare_dataset_ours_demo3} that when the noise level is high, the commercial software NI seems to employ a pre-defined pattern to smooth out noise and avoid artifacts.
\subsubsection{Discussion}
\begin{figure}[htbp]
\graphicspath{{Illus_image/Discussion/Crop/}}
\centering
\subfigure[Clean]{
\label{Fig4}
\includegraphics[width=1.06in]{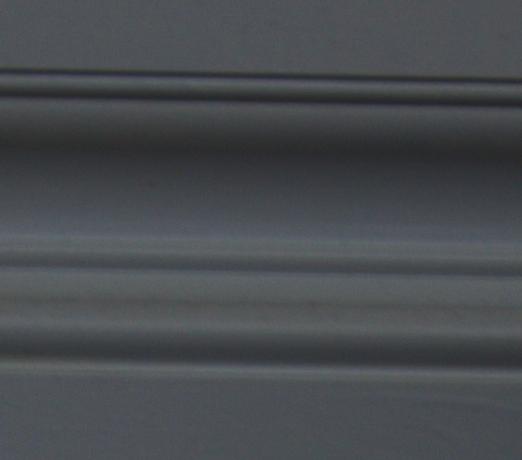}}
\subfigure[Noisy]{
\label{Fig4}
\includegraphics[width=1.06in]{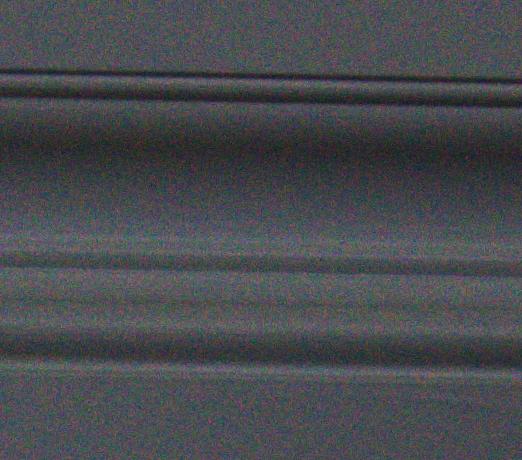}}
\subfigure[Reisized]{
\label{Fig4}
\includegraphics[width=0.94in]{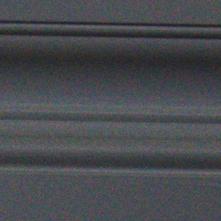}}\\

\subfigure[Clean]{
\label{Fig4}
\includegraphics[width=1in]{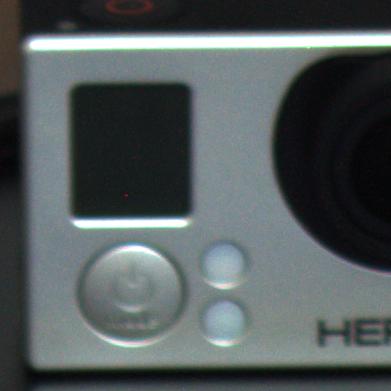}}
\subfigure[Noisy]{
\label{Fig4}
\includegraphics[width=1in]{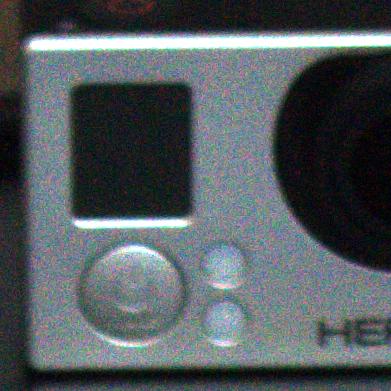}}
\subfigure[Reisized]{
\label{Fig4}
\includegraphics[width=1in]{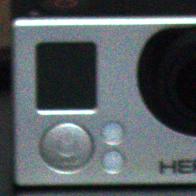}}

\caption{Visual comparison of clean, noisy and resized image of the same scene from RENOIR dataset. The camera is CANON T3i. The resized images are generated by MATLAB with 'Resized = imresize(Noisy,0.5)'. Please zoom-in for better view.}

\label{Fig_illus_resize}
\end{figure}

\begin{figure}[htbp]
\graphicspath{{Illus_image/Discussion/Combine/}}
\centering
\subfigure{
\label{Fig4}
\includegraphics[width=0.98in]{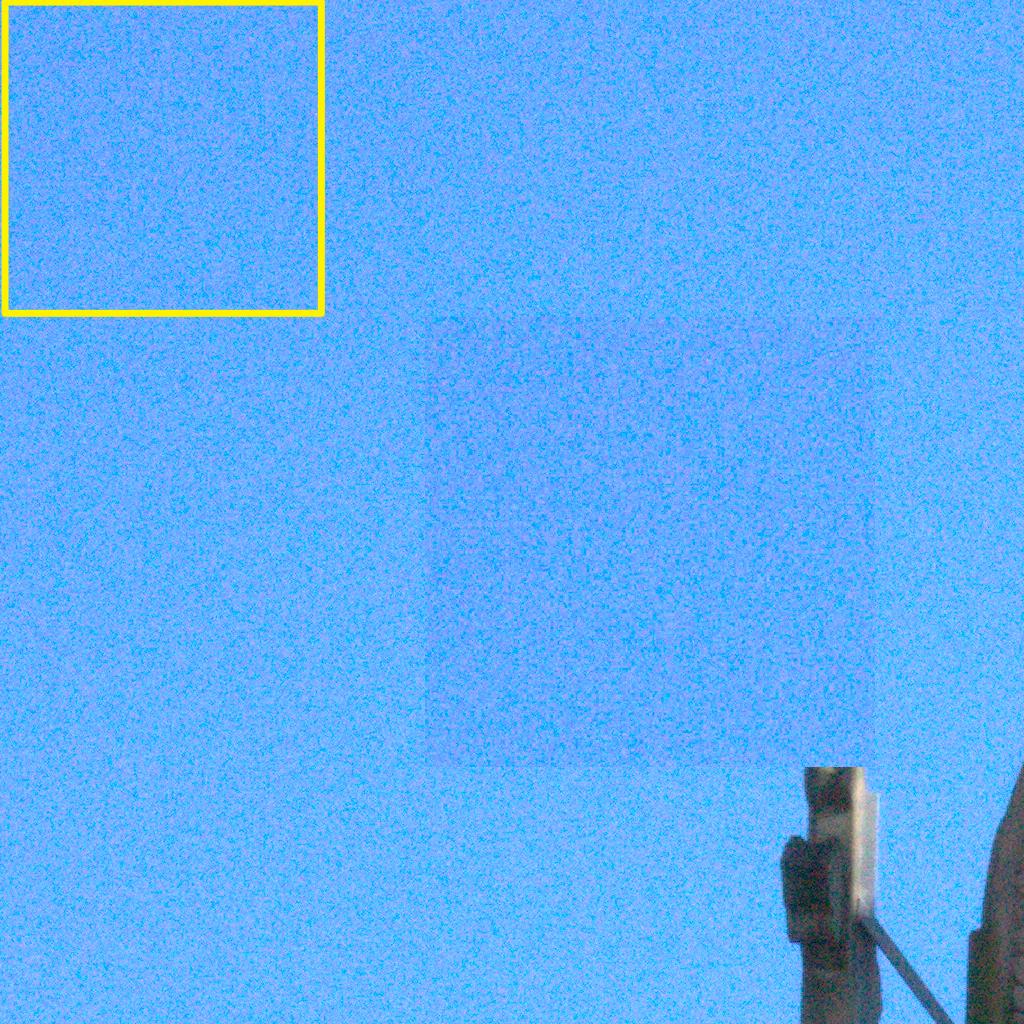}}
\subfigure{
\label{Fig4}
\includegraphics[width=0.98in]{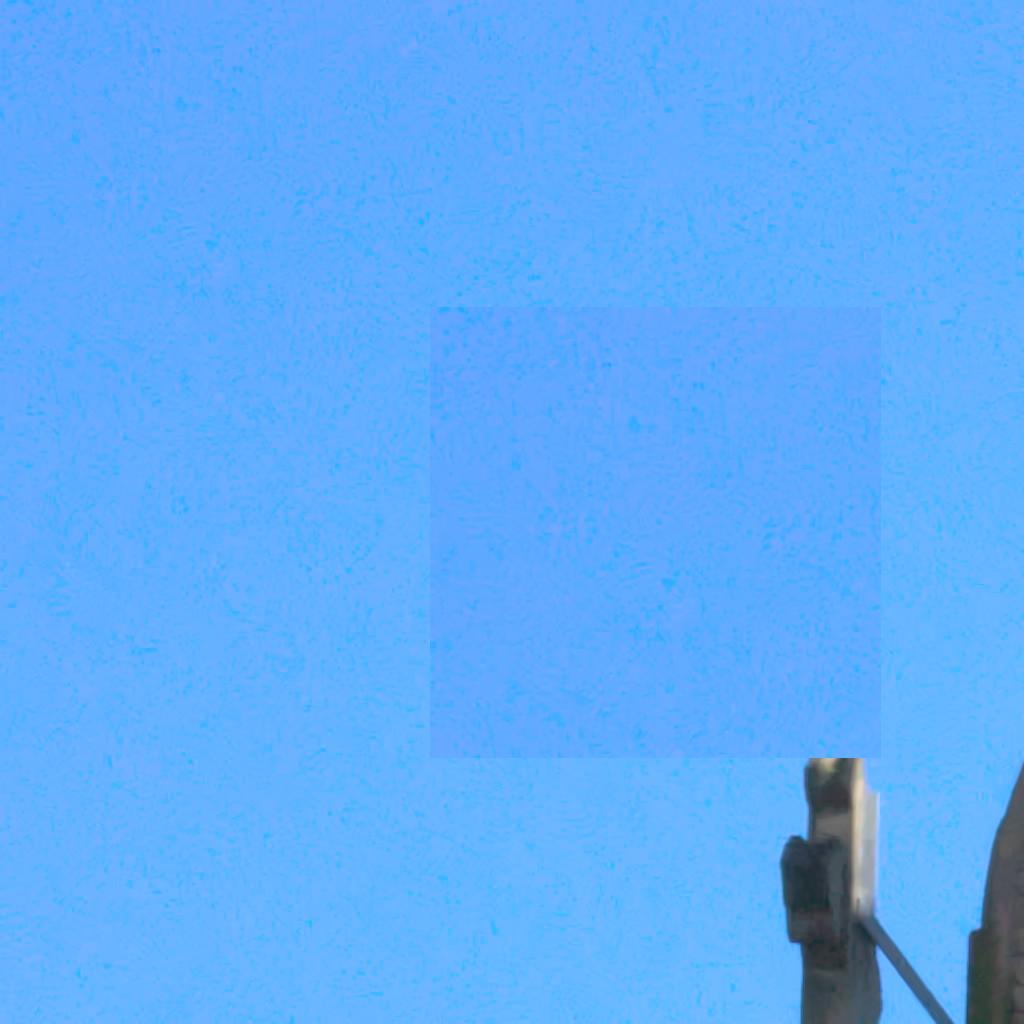}}
\subfigure{
\label{Fig4}
\includegraphics[width=0.98in]{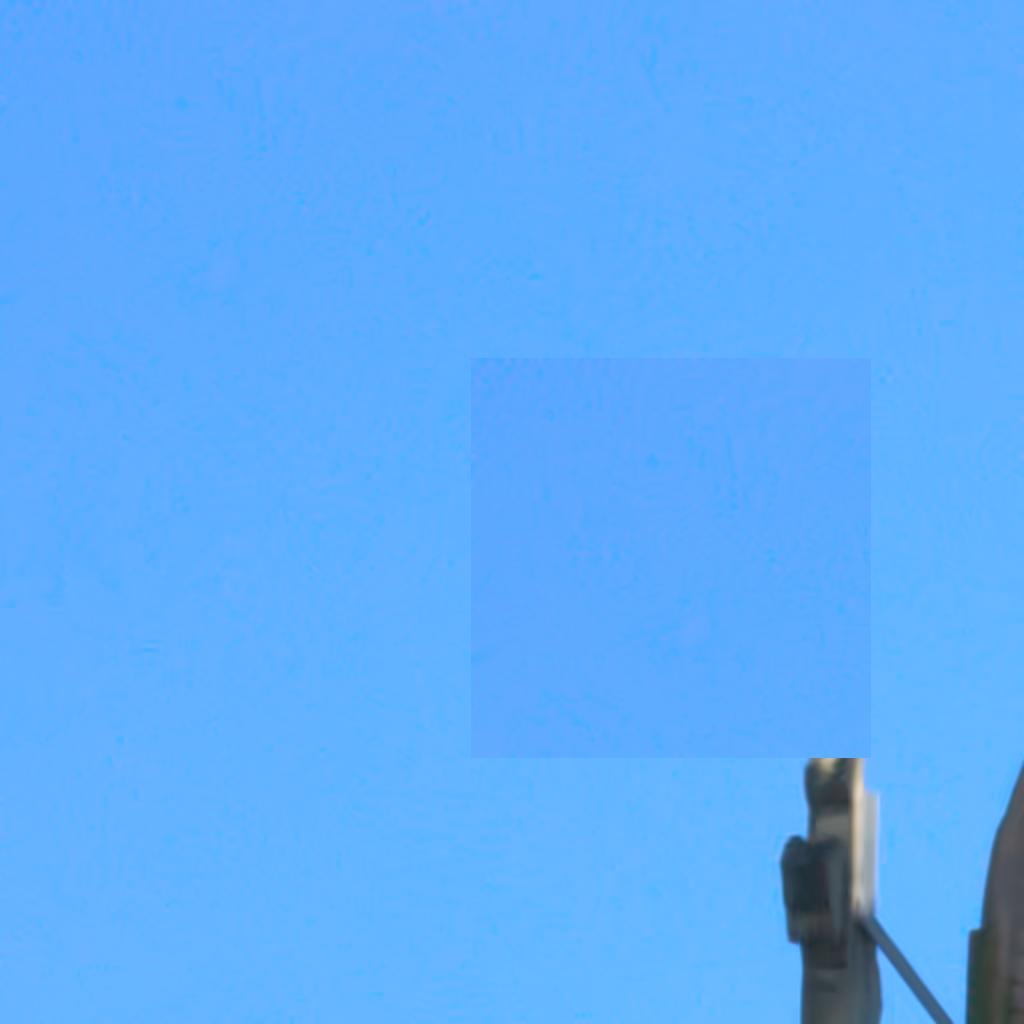}}

\subfigure{
\label{Fig4}
\includegraphics[width=0.98in]{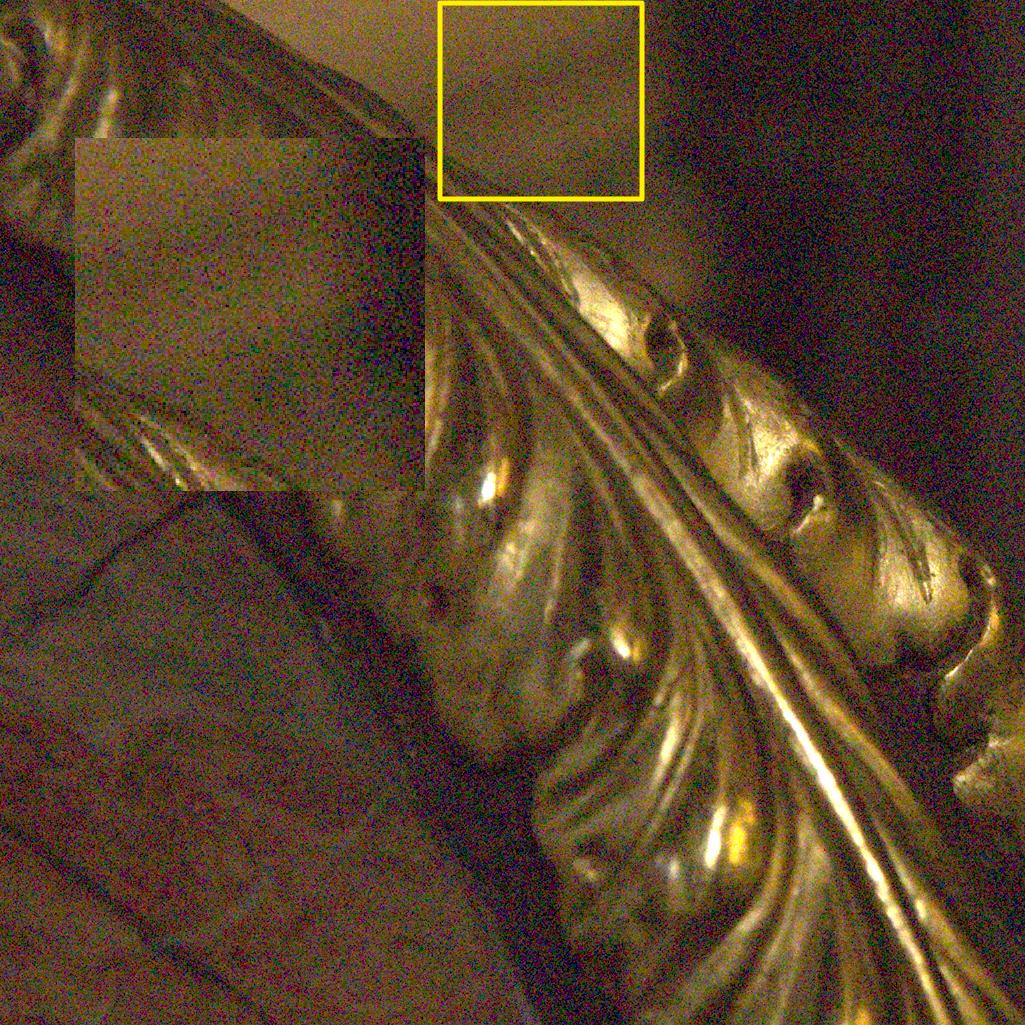}}
\subfigure{
\label{Fig4}
\includegraphics[width=0.98in]{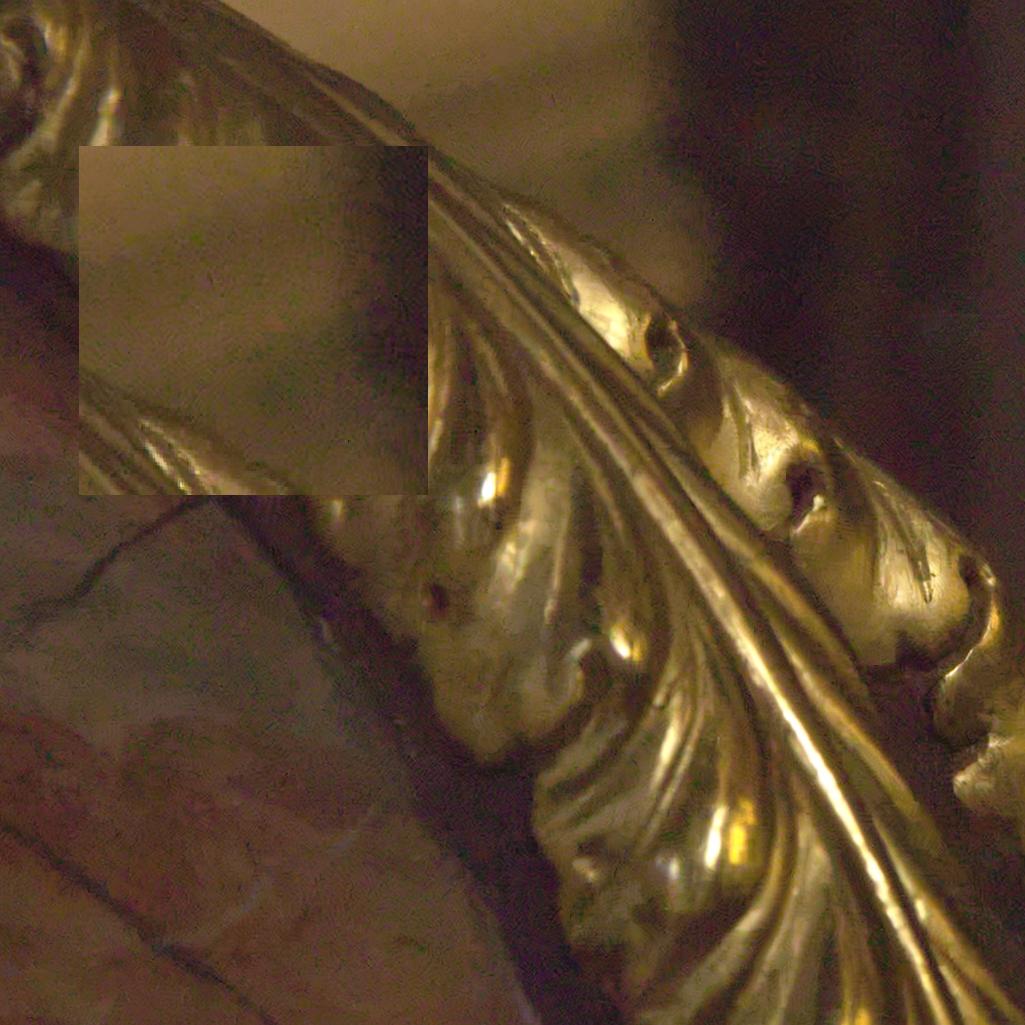}}
\subfigure{
\label{Fig4}
\includegraphics[width=0.98in]{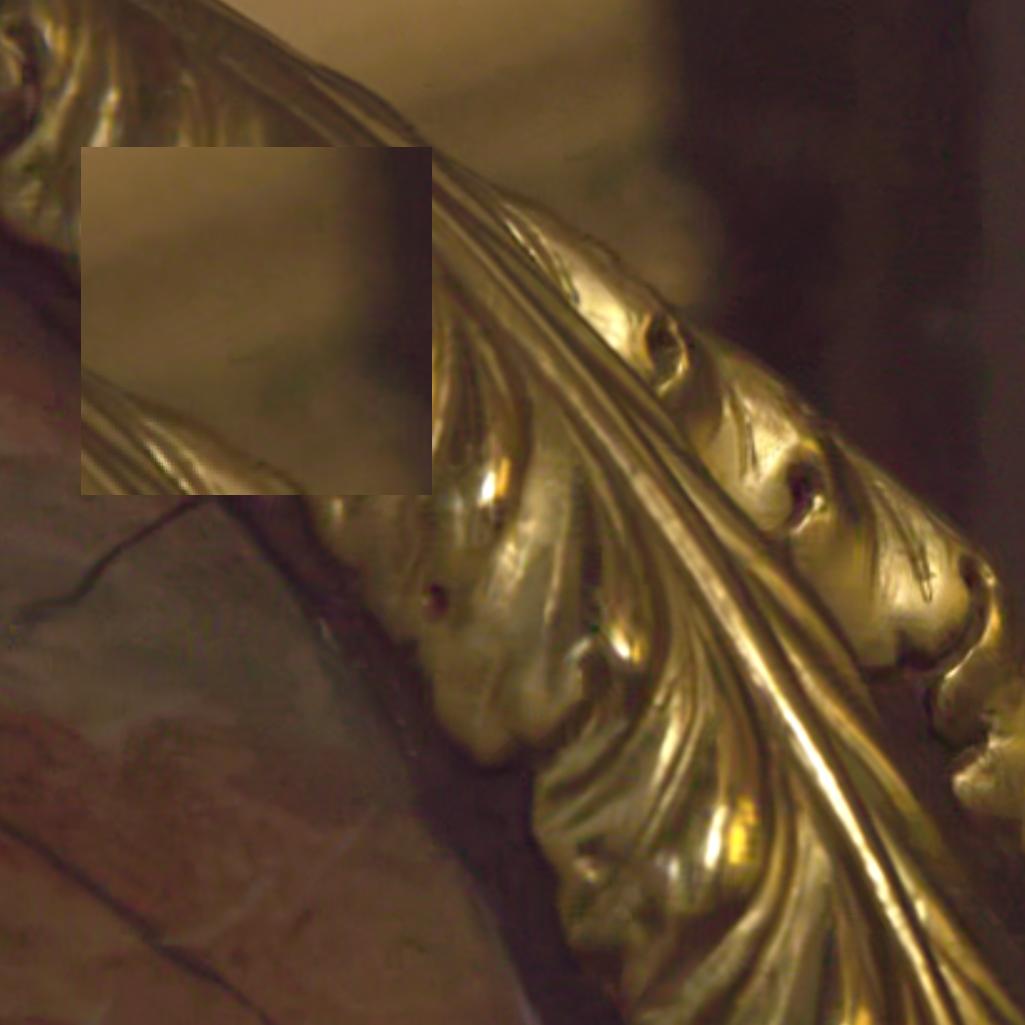}}

\subfigure{
\label{Fig4}
\includegraphics[width=0.98in]{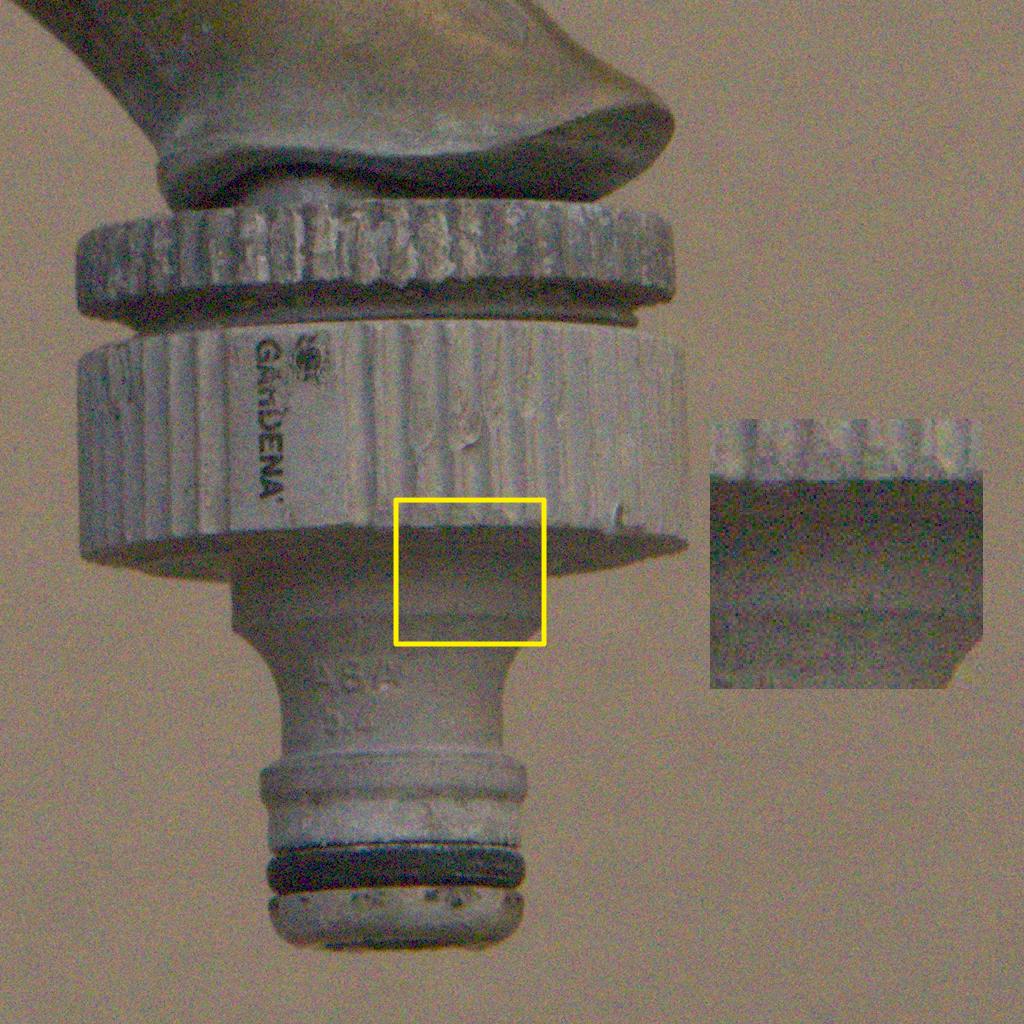}}
\subfigure{
\label{Fig4}
\includegraphics[width=0.98in]{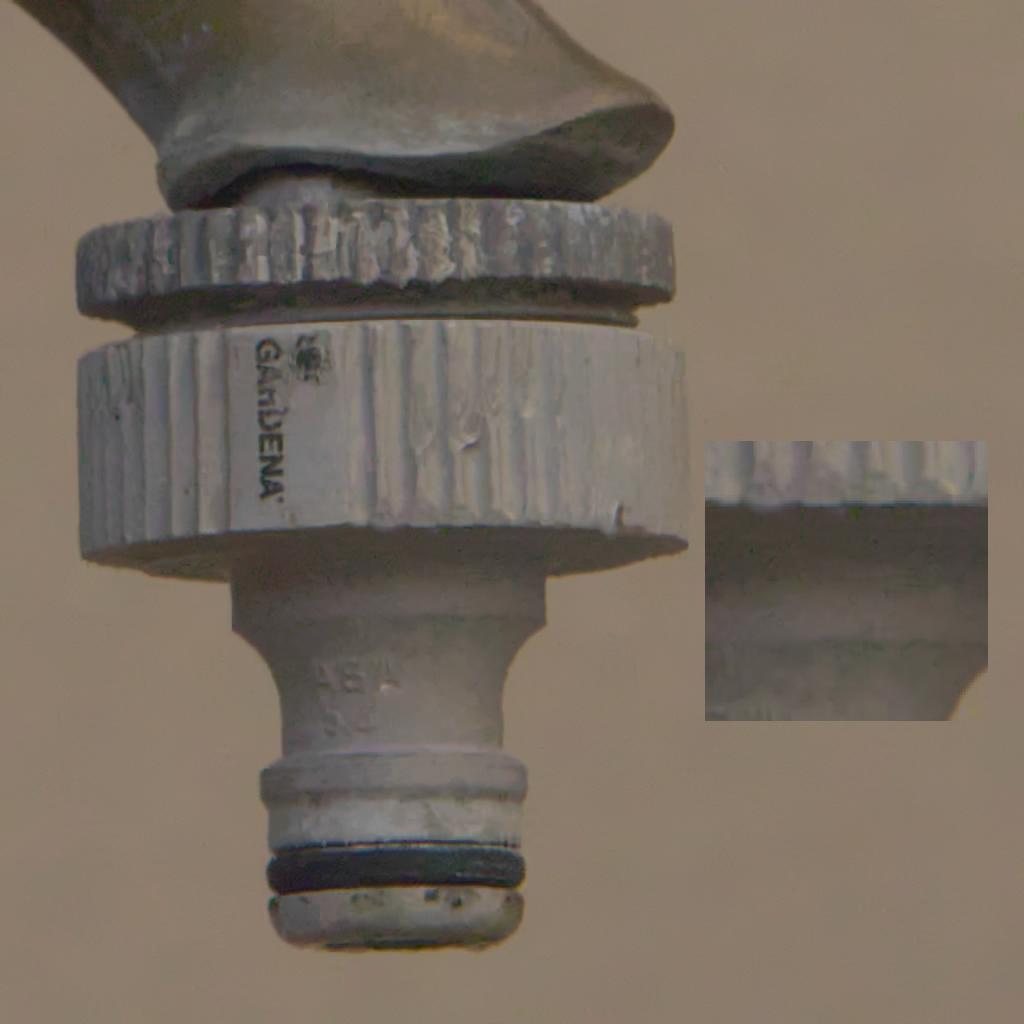}}
\subfigure{
\label{Fig4}
\includegraphics[width=0.98in]{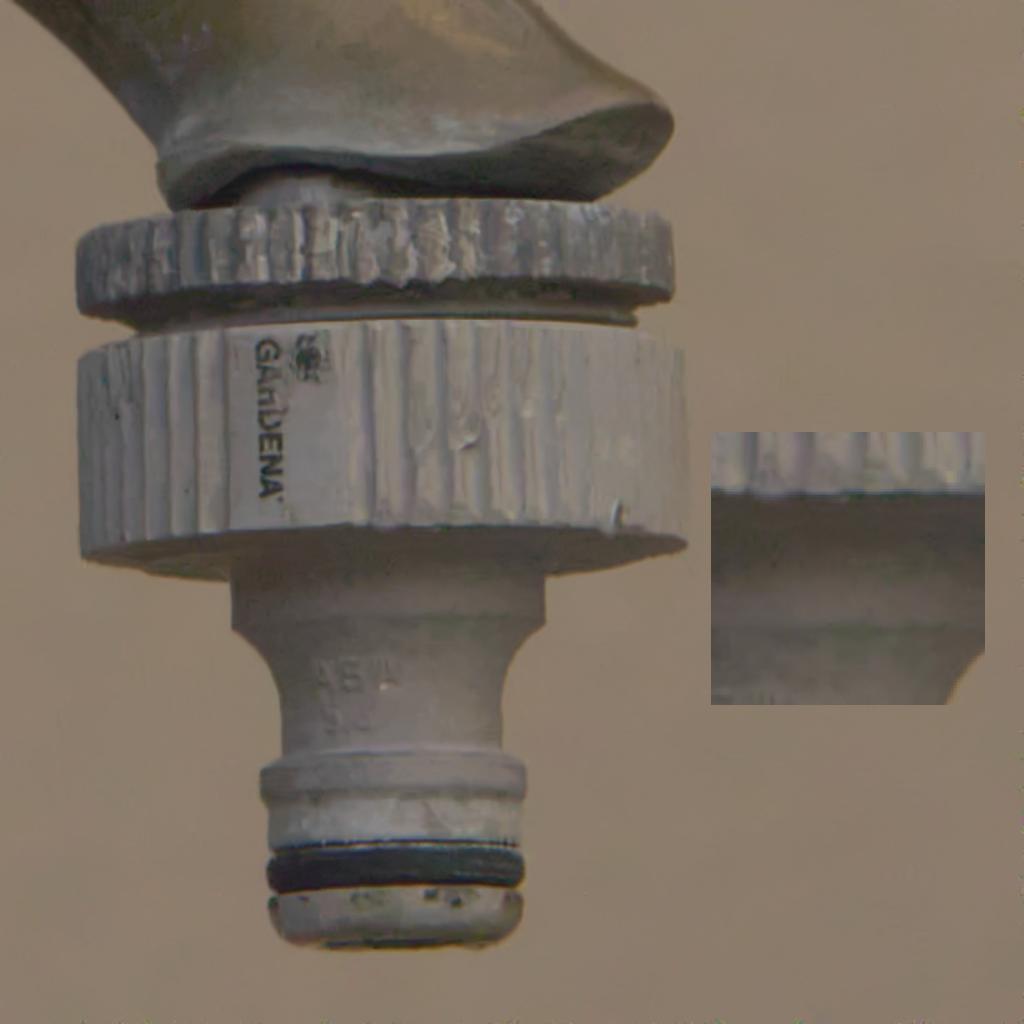}}

\caption{Visual evaluation of filtered images of DND dataset. From left to right: Noisy observation, MS-TSVD and resized MS-TSVD. Please zoom-in for better view.}

\label{Fig_illus_resize_denoise_dnd}
\end{figure}

\begin{figure}[htbp]
\graphicspath{{Illus_image/Discussion/Crop/}}
\centering
\subfigure{
\label{Fig4}
\includegraphics[width=0.98in]{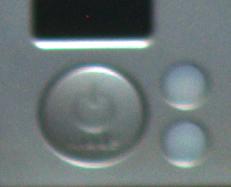}}
\subfigure{
\label{Fig4}
\includegraphics[width=0.98in]{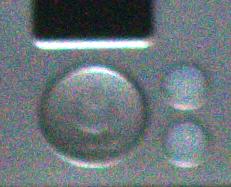}}
\subfigure{
\label{Fig4}
\includegraphics[width=0.98in]{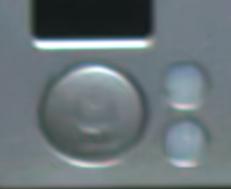}}

\subfigure{
\label{Fig4}
\includegraphics[width=0.98in]{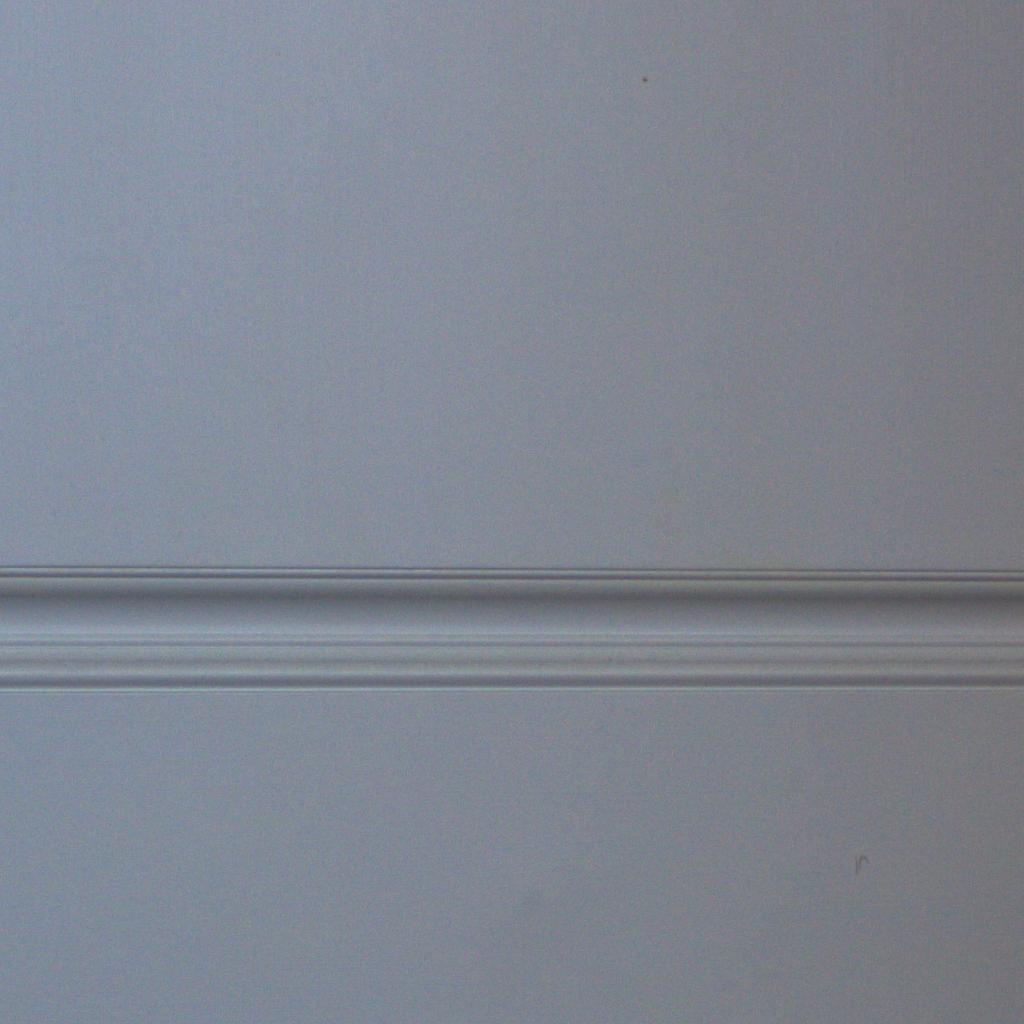}}
\subfigure{
\label{Fig4}
\includegraphics[width=0.98in]{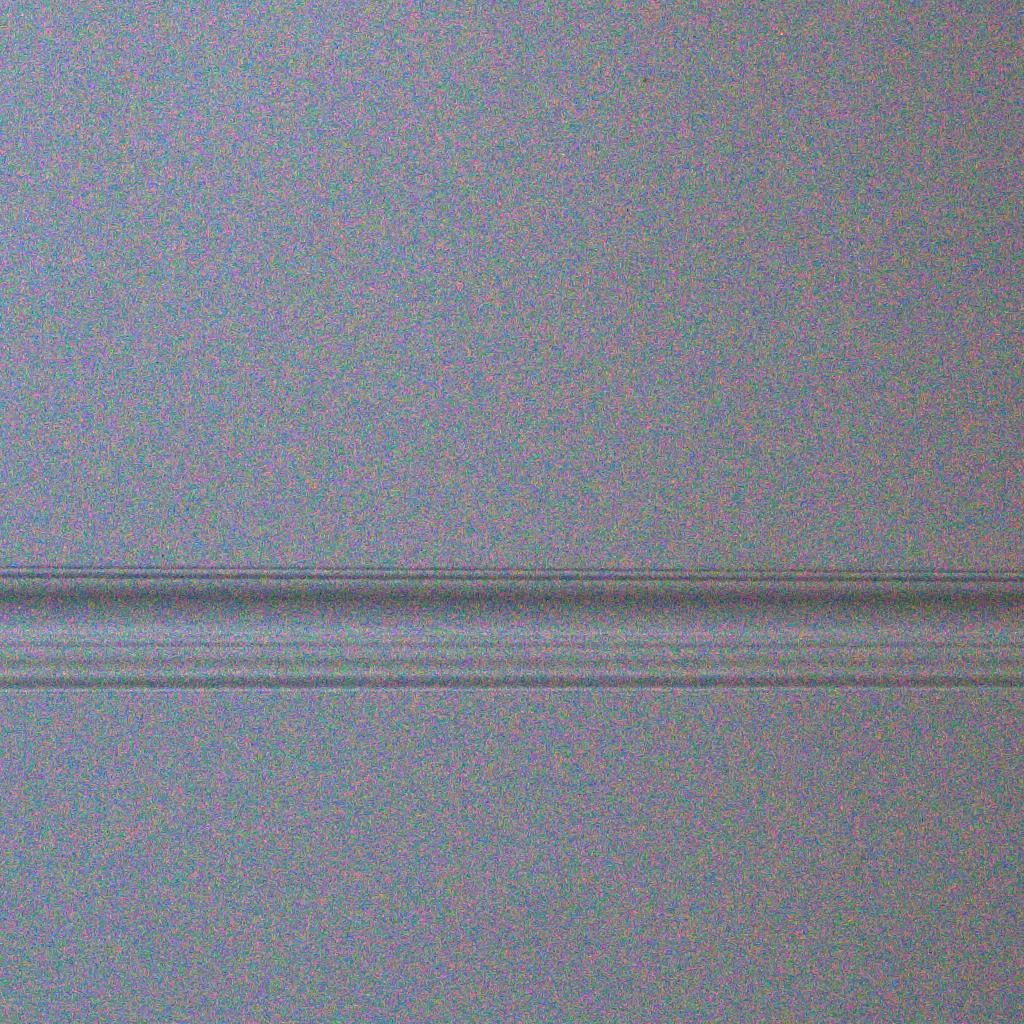}}
\subfigure{
\label{Fig4}
\includegraphics[width=0.98in]{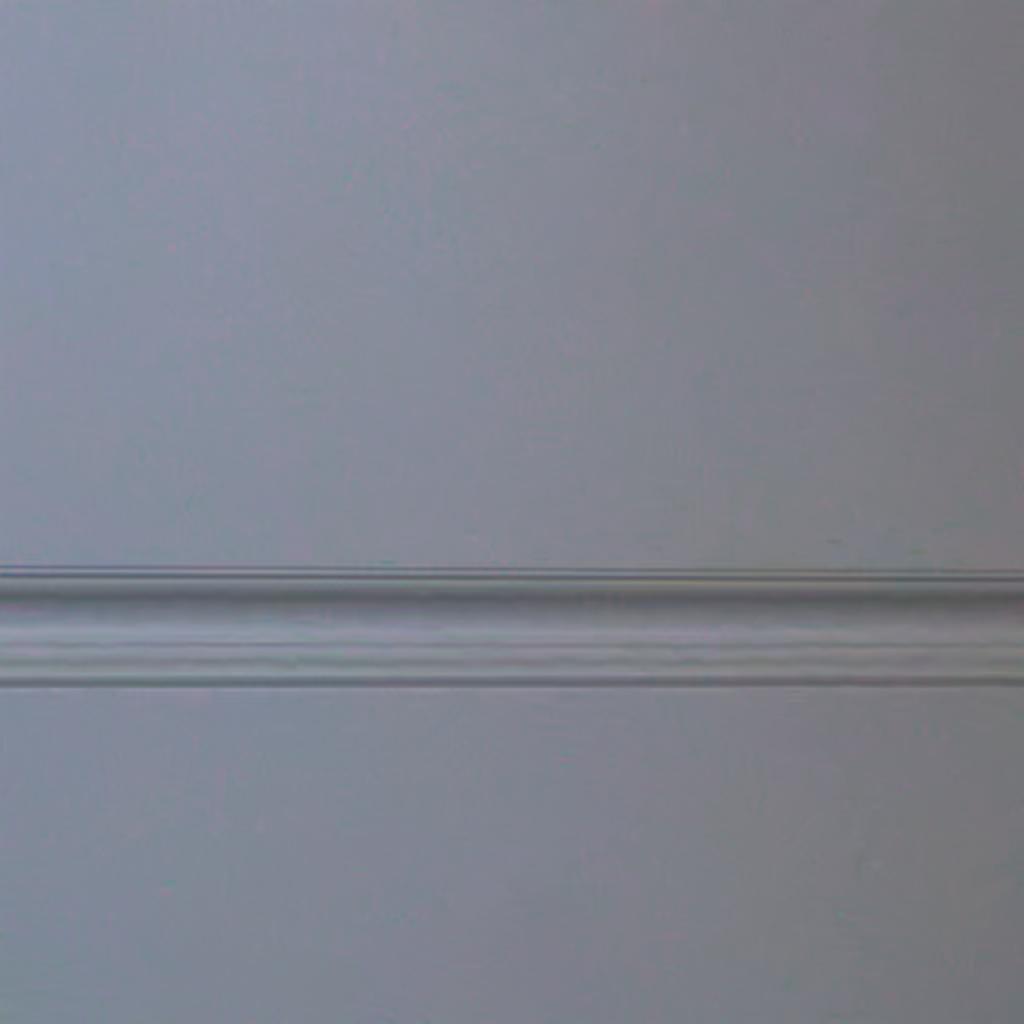}}

\caption{Visual evaluation of filtered images of RENOIR dataset. From left to right: Clean observation, Noisy observation, and resized MS-TSVD. Please zoom-in for better view.}

\label{Fig_illus_resize_denoise_renoir}
\end{figure}

Our comprehensive experiments show that CBM3D and MS-TSVD demonstrate the most competitive performance of all comparison methods. But Fig. \ref{Fig_compare_dataset1_demo1} shows that they also produce annoying artifacts on severely corrupted homogenous region, mainly due to the presence of strong noise in grouping and training steps. Although the strategy of NI risks sacrificing details, it shows satisfactory smooth visual effects. The implementation of NI is not publicly available, but similar to LLRT and TWSC, one plausible solution is to incorporate the low-rank approximation and sparse coding technique into MS-TSVD, however choosing proper ranks is not easy. Consider that in real cases, the parameters should be carefully tuned, an efficient and effective strategy should be utilized. In this subsection, we use some challenging images from RENOIR and DND datasets to demonstrate how to effectively produce smooth effects using current state-of-the-art framework.\\
 \indent Recently, \cite{Zontak2013Separating} builds a pyramid and shows that in the downsampled image of noisy observation, patches tend to be noiseless and share more similar pattern with original underlying clean patches than the corresponding noisy ones. This observation is illustrated in Fig. \ref{Fig_illus_resize} with images from RENOIR dataset. Therefore, instead of directly filtering noisy observation, an alternative is to first handle downsized image, and then upscale the denoised image back to its original size with some effective image super-resolution algorithms \cite{tang2017pairwise,dong2016image}. In this chapter, we use the simplest build-in bicubic function of MATLAB. Fig. \ref{Fig_illus_resize_denoise_dnd} and Fig. \ref{Fig_illus_resize_denoise_renoir} compare the visual effects of results produced by MS-TSVD with and without this resize strategy. The "ground-truth" images of DND dataset are not available, but the obvious smooth effects with less color and claw artifacts can be clearly seen. Another straightforward benefit of the resize strategy is efficiency, since the size of downsampled image is much smaller than that of the original one. But in real cases, this strategy should be used very carefully, because it is a tradeoff of details for smoothness.

\section{Conclusion}
In this paper, we present a brief review of real-world color image denoising framework and methodology. We describe several publicly available real-world color image datasets and introduce a newly constructed dataset for more comprehensive evaluation. Our experiments give an objective view on the effectiveness and efficiency of competing methods. Challenges and potentials of improving visual effects in real cases are also discussed.\\
\indent Future work includes incorporating External priors \cite{roth2009fields,luo2015adaptive} and new grouping strategy \cite{foi2016foveated} into current best methods.


\section*{Acknowledgment}
The authors would like to thank all authors of related methods for providing their code and software package.



%
\bibliographystyle{IEEEtran}
\bibliography{IEEEabrv,ms}

\end{document}